\pdfoutput=1
\PassOptionsToPackage{backref=page,breaklinks=true}{hyperref}
\documentclass[11pt]{article}

\newif\ifauthordecided
\authordecidedtrue

\newif\ifarxiv
\newif\ifcameraready

\arxivtrue        % true = arXiv submission version (use preprint)
\newif\ifperfect
\perfectfalse

\ifarxiv
    \usepackage[preprint]{acl} % arXiv version
\else
    \ifcameraready
        \usepackage[final]{acl} % final camera-ready version
    \else
        \usepackage[review]{acl} % review submission version
    \fi
\fi

\usepackage{times}
\usepackage{latexsym}

\usepackage[T1]{fontenc}
\usepackage[utf8]{inputenc}

\usepackage{microtype}

\usepackage{inconsolata} % to beautify the typewriter font \texttt

\usepackage{package} % Zhijing's package

\title{How Do Linear Probes Emerge? A Circuit-Tracing Framework with Concept-Targeted Attribution}
\ifauthordecided
\vspace{2em}
\author{
  \textbf{Vedant Palit}\textsuperscript{1,2,3} \quad
  \textbf{Florent Draye}\textsuperscript{3,4} \quad
  \textbf{Terry Jingchen Zhang}\textsuperscript{1,2} \\
  \textbf{Bernhard Sch\"olkopf}\textsuperscript{3,5} \quad
  \textbf{Zhijing Jin}\textsuperscript{1,2,3} \\[2mm]
  \textsuperscript{1}Jinesis Lab, University of Toronto \& Vector Institute \quad
  \textsuperscript{2}EuroSafeAI \\
  \textsuperscript{3}Max Planck Institute for Intelligent Systems, T\"ubingen, Germany \\
  \textsuperscript{4}Hector Foundation \quad
  \textsuperscript{5}ELLIS Institute T\"ubingen \\[2mm]
  \texttt{\{vedantpalit,zjin.admin\}@cs.toronto.edu}
}
\fi

\begin{document}
\addtocontents{toc}{\protect\setcounter{tocdepth}{-1}}
\maketitle

\begin{abstract}
Transcoder attribution graphs are usually trained to explain why a model assigns high probability to a particular next token. We introduce \textbf{Concept-Targeted Attribution} (CTA), which instead trains attribution graphs with respect to a linear probe direction. CTA therefore yields probe-specific circuits that explain why an internal concept representation arises in a prompt, independently of whether it is expressed in the generated token. Using Cross-Layer Transcoders, we show that these probe-targeted graphs contain predictive structure: graph-level features predict probe accuracy across four widely studied concept categories ($\rho = 0.91$, $R^2 = 0.84$), while local features identify the sparse components driving per-prompt classification. This connects probe performance to interpretable circuit structure, allowing us to ask not only whether a probe works, but which internal computations make it work. Causal ablations further show that probe-targeted and logit-targeted graphs capture functionally distinct mechanisms. Removing probe-relevant features reduces internal concept scores while largely preserving generated tokens, whereas removing logit-relevant features changes the generated token in $92\%$ to $100\%$ of cases with near-zero effect on probe scores. CTA provides a framework for moving from behavioral probe accuracy to mechanistic explanations of probe performance, enabling more detailed audits of internal concept representations, including safety-critical ones.\footnote{Our code is  at \url{https://github.com/vedantpalit/concept-targeted-attribution}.}
\end{abstract}

\section{Introduction}
\label{section:introduction}

\begin{figure*}[t]
    \centering
    \includegraphics[width=\linewidth, trim={20pt 30pt 0 30pt}, clip]{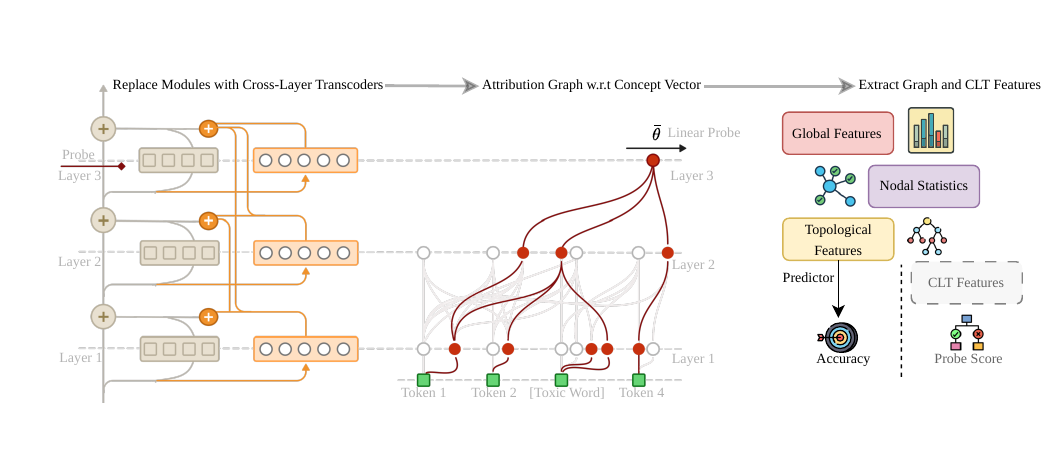}
    \caption{The CTA pipeline: CLT replaces MLP modules for interpretability; attribution graphs capture causal feature flow toward the probe; graph-structural features predict probe accuracy; CLT feature activations predict the probe margin.}
    \label{fig:anchor_figure}
\end{figure*}

Mechanistic interpretability has made remarkable progress through two complementary approaches \citep{bereska2024mechanistic}. \textbf{Feature decomposition} uses sparse recovery methods to decompose polysemantic residual streams into dictionaries of interpretable features \citep{huben2024sparse}, while \textbf{circuit localization} identifies the model components responsible for a given behavior \citep{conmy2023automatedcircuitdiscoverymechanistic}. Recent attribution-graph methods combine these ideas by tracing feature-level influence through the model toward a final next-token logit~\citep{ameisen2025circuit}. This has made it possible to study token generation at the level of sparse computational structure \citep{lindsey2025biology}.

However, grounding every attribution graph in a next-token logit creates a \textit{token bottleneck}. A model may internally represent a concept without expressing it in the token it generates. For example, when processing a toxic quote in a neutral framing, such as \textit{The harassment report included: ``[Toxic Quote].'' The platform...}, the model may safely predict \textit{removed}. A logit-targeted graph can explain why the model produces this safe continuation, but it does not directly explain how the toxic content was recognized and represented internally.

Linear probes~\citep{Alain2017Understanding} offer a complementary view by testing whether a concept is decodable from the model's hidden activations. Yet probes are usually diagnostic rather than mechanistic: they tell us whether a concept is linearly accessible, but not which sparse features support the detection, why probe performance varies across layers or prompts, or how concept detection relates to token generation.

We introduce \textbf{Concept-Targeted Attribution (CTA)}, a method for constructing attribution graphs with respect to internal concept representations rather than output tokens. Given a linear probe trained to detect a concept in hidden activations, CTA uses the probe score as the scalar target for attribution. The resulting graph answers: ``Which internal computations made this concept detectable in the model for this prompt?'' In this way, CTA turns probes from post-hoc diagnostic tools into targets for circuit-level analysis.

We apply CTA to four widely studied probes: toxicity, sentiment, reasoning, and truthfulness. Our analysis addresses three questions. First, are probe-targeted attribution graphs functionally distinct from logit-targeted graphs? We find a causal double dissociation: ablating probe-exclusive features reduces concept scores while largely preserving the generated token, whereas ablating logit-exclusive features changes the generated token in $92\%$--$100\%$ of cases with near-zero effect on concept scores. Second, can attribution-graph structure explain probe quality? Graph-level features predict probe validation accuracy across the four concepts ($\rho = 0.91$, $R^2 = 0.84$), while local features identify the sparse latents that drive per-prompt classification. Third, how consistent are probe-targeted graphs across prompts and layers?  We find that while attribution graphs diverge at the periphery, they converge on a shared feature core that is recruited early and persists through the network, consistent with multiple distinct circuits converging on the same concept representation~\citep{chen2026circuitsleadromerethinking}.

\section{Related Work}
\label{section:related work}
\noindent \textbf{Circuit Discovery and Attribution Graphs}
The dominant circuit discovery paradigm begins with activation patching, which tests whether replacing an activation in one run with the corresponding activation from another run changes the model's output~\citep{vig2020causalmediationanalysisinterpreting, meng2022neuripslocating, wang2023interpretability}. This identifies components whose activations are causally important for a behavior. Automated methods extend this idea by searching over many components to recover larger computational flow graphs through the model~\citep{conmy2023automatedcircuitdiscoverymechanistic, hanna-etal-2025-circuit}.

Dictionary-learning methods provide a more fine-grained unit of analysis. Sparse Autoencoders decompose model activations into overcomplete sets of sparse latent features, so that a residual-stream vector can be represented as a small combination of more interpretable directions~\citep{bricken2023monosemanticity, templeton2024scaling}. Transcoders apply the same idea to model computations: instead of reconstructing an activation vector, they learn sparse features that reconstruct the output of an MLP block \citep{dunefsky2024transcoders}. Cross-layer transcoders extend this setup across layers, allowing one to track how sparse features contribute to later residual updates. Both methods make it possible to build feature-level attribution graphs, where influence is traced through sparse latents and their linear paths rather than through coarse components such as neurons, heads, or layers~\citep{ameisen2025circuit}.

In current attribution-graph work, however, the target of this tracing procedure is typically an output logit. The resulting graph explains which sparse features and paths increased the probability of a particular next token. This logit-centric framing is powerful for studying token generation, but it does not directly describe internal concept representations that may be present without being expressed in the generated token.

\noindent \textbf{Concept Directions and Diagnostic Probing}
A parallel line of work studies concepts as directions in activation space. Linear probes are simple classifiers trained on hidden activations: if a probe can predict a concept label from a layer's residual stream, the concept is taken to be linearly decodable at that layer~\citep{Alain2017Understanding, burns2023discovering, zou2023representation, marks2024geometrytruthemergentlinear}. Representation-engineering methods similarly identify directions associated with high-level properties and use them to analyze or steer model behavior~\citep{zou2023representation,marks2024geometrytruthemergentlinear}. Concept-based explanation methods such as TCAV also represent concepts as activation-space directions, then measure whether moving along those directions affects model predictions~\citep{pmlr-v80-kim18d}.

These methods establish that high-level concepts can often be read out from hidden states, but they do not by themselves give a circuit-level explanation of the readout. A probe can tell us that toxicity, sentiment, reasoning, or truthfulness is decodable from a layer, but not which sparse features make the probe fire, how those features are assembled across the computation, or how the concept-level signal relates to the paths used for next-token generation. 
%Even if accuracies are same, do the features overlap and which to use?
%Study the graphs for the examples which fail

\section{Methodology}
\label{section:methods}

% To evaluate the structural and functional properties of internal concept representations, we introduce Concept-Targeted Attribution (CTA). We first define the baseline mechanisms of our feature space and attribution targets, followed by the specific experimental designs used to test circuit dissociation and topological 
% predictability.

\subsection{Transcoder Attribution Graphs}
\label{subsection:preliminaries}

Transcoder attribution graphs replace the MLP, the main nonlinear part of a transformer block, with a sparse feature model. Given the residual stream $x_\ell$ entering layer $\ell$, a transcoder reconstructs the MLP output as
\begin{align}
    \mathrm{MLP}_\ell(x_\ell) \approx \sum_i f_{\ell,i} d_{\ell,i} + b_\ell ,
\end{align}
where $f_{\ell,i}$ are sparse feature activations and $d_{\ell,i}$ their decoder directions. This makes attribution tractable: once the MLP is replaced by sparse feature contributions, attribution can be followed through the model's linear paths. The resulting graph has sparse features as nodes and attribution weights as edges.

An attribution graph explains a scalar target. In standard logit-targeted attribution, this target is a next-token logit $z_y$, so the graph traces which sparse feature paths made the model more likely to generate token $y$. In \textbf{Concept-Targeted Attribution} (CTA), we keep the same graph construction but change the target: instead of explaining a token logit, we explain the score of a linear concept probe. For a concept direction $\bar{\theta} \in \mathbb{R}^d$ and residual stream activation $x_\ell$, the probe score is
$S_{\bar{\theta}}(x_\ell) = \bar{\theta}^{\top} x_\ell$.
This score measures how strongly the residual stream aligns with the concept direction at a specific token position. It is a prompt-level scalar, distinct from probe accuracy, which is a dataset-level classification metric over held-out examples.
\begin{figure}[t]
    \centering
    \includegraphics[width=0.8\linewidth]{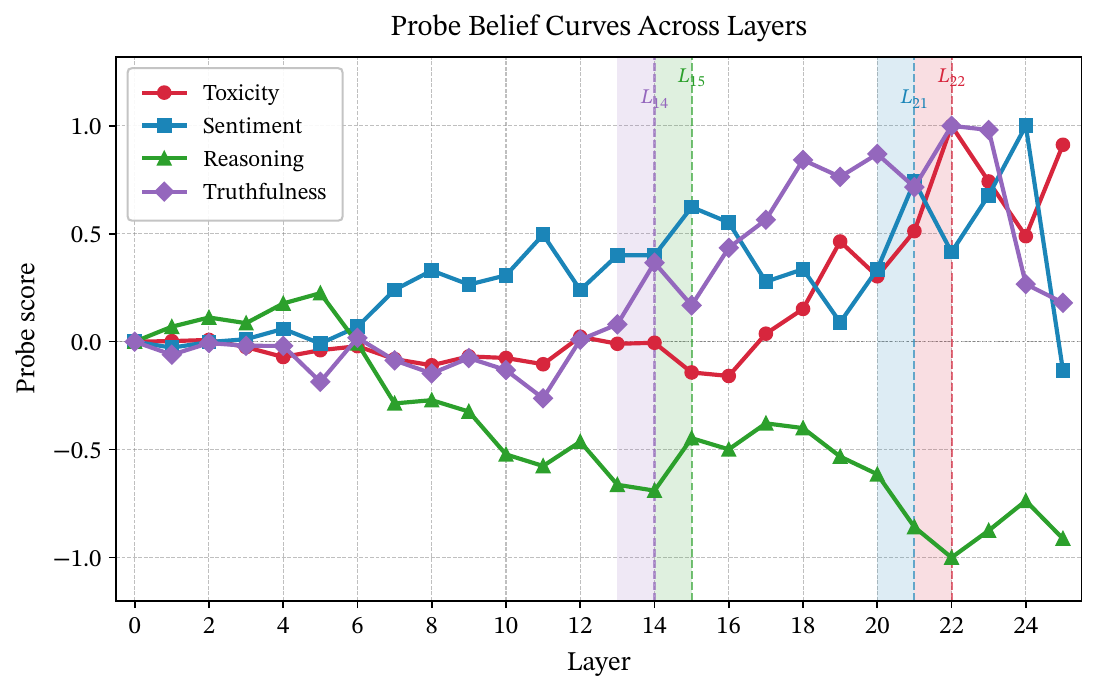}
    \caption{Probe score trajectories across layers for all four concepts. Shaded bands mark the steepest single-step rise, indicating the belief point $L_b$ where the residual stream most sharply aligns with the concept direction.}
    \label{fig:probe_overview}
\end{figure}
A CTA graph therefore traces which sparse features and paths make a concept detectable in the residual stream for a particular prompt. In our experiments, we use Cross-Layer Transcoders and track changes in the probe score $\Delta S_{\bar{\theta}}$ for causal dissociation experiments; further implementation details are provided in Appendix~\ref{appendix: experimentation details}.
\subsection{Concept Representation via Probing}
\label{subsection:concept_probing}

We evaluate the model across four semantic concepts spanning a range of the transformer's internal computational lifecycle: (i) \textit{Toxicity} \citep{logacheva-etal-2022-paradetox}, a safety-critical intent filter isolated from surface keywords via paired semantic contexts (ii) \textit{Sentiment} \citep{socher-etal-2013-recursive}, an early-layer affective signal (iii) \textit{Reasoning} \citep{hong-etal-2025-reasoning}, contrasting multi-hop relational binding~\citep{yang-etal-2018-hotpotqa} against single-hop recall~\citep{joshi-etal-2017-triviaqa} and (iv) \textit{Truthfulness} \citep{marks2024geometrytruthemergentlinear}, characterising factual verification. We train linear classifiers over the residual stream at each layer and apply Gram-Schmidt orthogonalisation, projecting each concept direction onto the orthogonal complement of the preceding directions. Post-correction concept directions are near-orthogonal (maximum $|\cos(\bar{\theta})| = 0.069$), confirming that orthogonality is a property of the model's representational geometry rather than an artefact of post-processing. Full details of dataset curation, balancing, and training are provided in Appendix~\ref{appendix: experimentation details}.
\begin{table*}[t]
\centering
\small
\setlength{\tabcolsep}{4pt}
\resizebox{\textwidth}{!}{%
\begin{tabular}{l ccc ccc @{\hspace{1.5em}} ccc ccc}
\toprule
& \multicolumn{6}{c}{\textbf{Gemma-2-2B}}
& \multicolumn{6}{c}{\textbf{Llama-3.2-1B}} \\
\cmidrule(lr){2-7}\cmidrule(lr){8-13}
& \multicolumn{3}{c}{\textbf{Toxicity} \textit{(L22)}}
& \multicolumn{3}{c}{\textbf{Truthfulness} \textit{(L14)}}
& \multicolumn{3}{c}{\textbf{Toxicity} \textit{(L14)}}
& \multicolumn{3}{c}{\textbf{Truthfulness} \textit{(L1)}} \\
\cmidrule(lr){2-4}\cmidrule(lr){5-7}
\cmidrule(lr){8-10}\cmidrule(lr){11-13}
\textbf{Subspace}
  & \textbf{Post-}$S$ & $\boldsymbol{\Delta S}$ & \textbf{Flip}
  & \textbf{Post-}$S$ & $\boldsymbol{\Delta S}$ & \textbf{Flip}
  & \textbf{Post-}$S$ & $\boldsymbol{\Delta S}$ & \textbf{Flip}
  & \textbf{Post-}$S$ & $\boldsymbol{\Delta S}$ & \textbf{Flip} \\
\midrule
Clean
  & 102.52 & —      & 0.0\%
  & 41.87  & —      & 0.0\%
  & 0.43   & —      & 0.0\%
  & 0.11   & —      & 0.0\% \\
$G_{\text{probe}} \setminus G_{\text{logit}}$
  & 41.28   & \textcolor{blue}{\textbf{$-$61.24}} & 0.0\%
  & 29.84   & \textcolor{blue}{\textbf{$-$12.03}} & 25.0\%
  & $-$0.10 & \textcolor{blue}{\textbf{$-$0.53}}  & 0.0\%
  & 0.11    & $\approx$0                          & 0.0\% \\
$G_{\text{logit}} \setminus G_{\text{probe}}$
  & 101.82 & 0.70       & \textcolor{red}{\textbf{100.0\%}}
  & 41.84  & 0.03       & \textcolor{red}{\textbf{100.0\%}}
  & 0.43   & $\approx$0 & \textcolor{red}{\textbf{100.0\%}}
  & 0.09   & $-$0.02    & \textcolor{red}{\textbf{100.0\%}} \\
$G_{\text{probe}} \cap G_{\text{logit}}$
  & $+$36.50 & $-$66.02 & 38.0\%
  & $+$2.93  & $-$38.94 & 67.0\%
  & $+$0.67  & $+$0.24  & 78.0\%
  & $+$0.08  & $-$0.03  & 17.0\% \\
\bottomrule
\end{tabular}%
}
\caption{Causal dissociation for Toxicity and Truthfulness depicted at their respective $L_b$, implemented across Gemma-2-2B and Llama-3.2-1B. Ablating
probe-exclusive nodes collapses $\Delta S$ with near-zero behavioural flip while ablating logit-exclusive nodes flips generation in 100\% of cases with negligible impact on $S$. $\Delta S$ magnitudes differ between models due to scale, however the dissociation effects remain consistent across both. Extended results in
Appendices~\ref{appendix: detailed results} and \ref{app:llama}.}
\label{table:causal_dissociation}
\end{table*}
We define the \textbf{belief point} $L_b$ as the layer exhibiting the greatest single-step rise in mean positive-class probe score, marking the layer at which the residual stream most sharply aligns with the concept direction. All single-layer graph extractions are anchored to the belief point layer.

\subsection{Causal Dissociation}
\label{subsection:dissociation_design}

To test whether concept representation and token generation rely on distinct pathways, we use a causal dissociation analysis, following the classical logic of separating functional mechanisms through selective interventions~\citep{teuber1955physiological}. For each prompt, we construct two graphs: a \textbf{CTA graph} rooted at the concept probe direction $\bar{\theta}$, and an \textbf{LTA graph} rooted at the next-token logit. We then ablate graph-specific features. If CTA-exclusive features support concept representation, their removal should reduce the probe score $S_{\bar{\theta}}$ while preserving the generated token. If LTA-exclusive features support token generation, their removal should change the top-1 prediction while leaving $S_{\bar{\theta}}$ largely unchanged. For each prompt, we extract both graphs at the concept-specific belief point $L_b$ and evaluate four conditions: \textit{clean}, \textit{CTA-exclusive}, \textit{LTA-exclusive}, and \textit{intersection}. We report $\Delta S_{\bar{\theta}}$ and the top-1 logit flip rate.

\subsection{Predicting the Quality of a Probe}
\label{subsection:topology_predictor}
\noindent \textbf{Graph-Structural Predictor.} To evaluate whether the structural shape of a CTA graph predicts probe quality, we extract CTA graphs across all layers for all four concepts and compute 35 structural features per graph across three families: global metrics, node statistics, and topological features. Since graph structure is a property of the probe direction rather than individual prompts, features are averaged across all prompts per (concept, layer) pair, yielding one stable feature vector per pair. The objective is to predict probe validation accuracy at layer $L$ using only these structural features. We train a Gradient Boosting regressor as well as a Ridge regressor under 5-fold group-aware cross-validation, splitting strictly by (concept, layer) to prevent target leakage, and evaluate using Spearman's $\rho$ as the primary metric. We also include a layer-mean baseline that predicts each layer's mean validation accuracy across concepts, as a sanity check against layer-depth confounding.

\noindent \textbf{Local Feature Predictor.} As a fine-grained analysis, we examine whether transcoder features active in the probe attribution graphs can predict per-prompt classification confidence. For each concept, we use the  features of the transcoder by activation strength as regressors and predict the probe score per prompt via Ridge regression with 5-fold cross-validation. We additionally identify \textit{rise features}, which are features significantly enriched in prompts that transition from incorrect to correct at the belief point (Fisher's exact test, $p < 0.05$) in order to pinpoint the specific sparse latents that drive individual probe decisions. Full model parameters, feature extraction details and validation protocols are provided in Appendix~\ref{appendix: experimentation details}.
%\section{Problem Formulation}
%\section{(Optional) Dataset Collection}
%\section{Method}
\section{Experiments}
\label{section:experiments}

Unless otherwise stated, the main paper reports results for Gemma-2-2B. Results for Llama-3.2-1B are provided in Appendix \ref{app:llama}.

\subsection{Concept Emergence and the Belief Point}
\label{subsection:belief_point}

\begin{figure*}[t]
\centering
\begin{minipage}{0.2\textwidth}
    \centering
    \includegraphics[width=\linewidth]{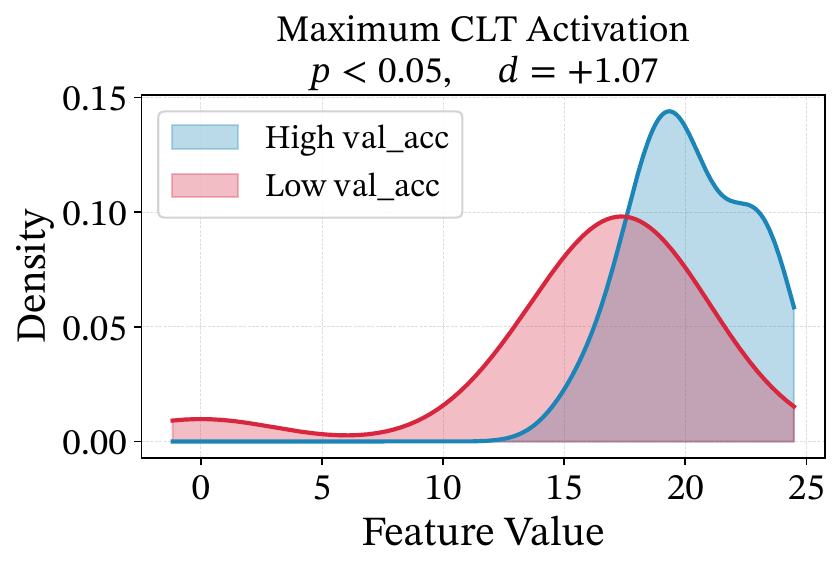}
\end{minipage}\hfill
\begin{minipage}{0.2\textwidth}
    \centering
    \includegraphics[width=\linewidth]{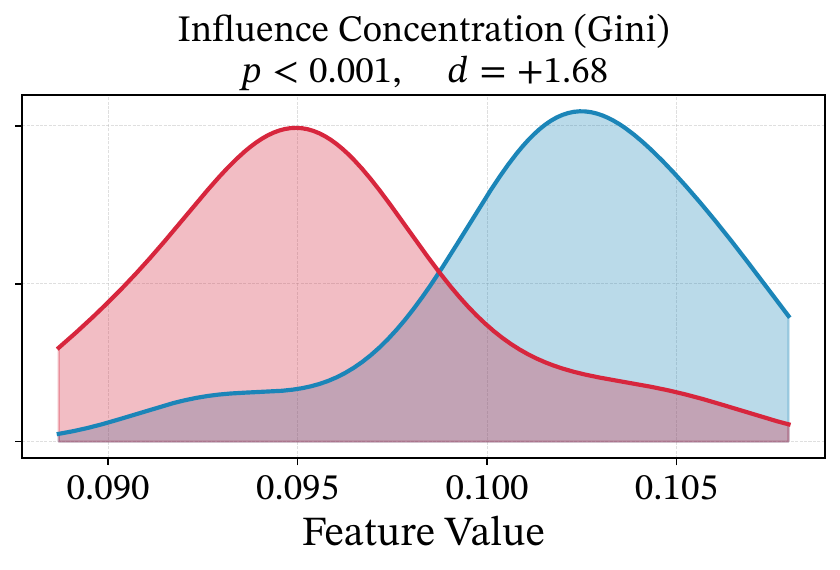}
\end{minipage}\hfill
\begin{minipage}{0.2\textwidth}
    \centering
    \includegraphics[width=\linewidth]{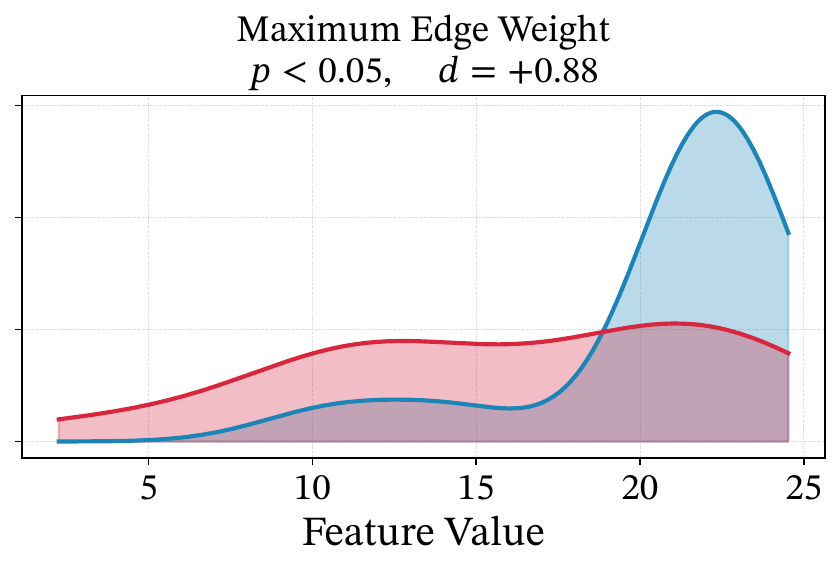}
\end{minipage}\hfill
\begin{minipage}{0.2\textwidth}
    \centering
    \includegraphics[width=\linewidth]{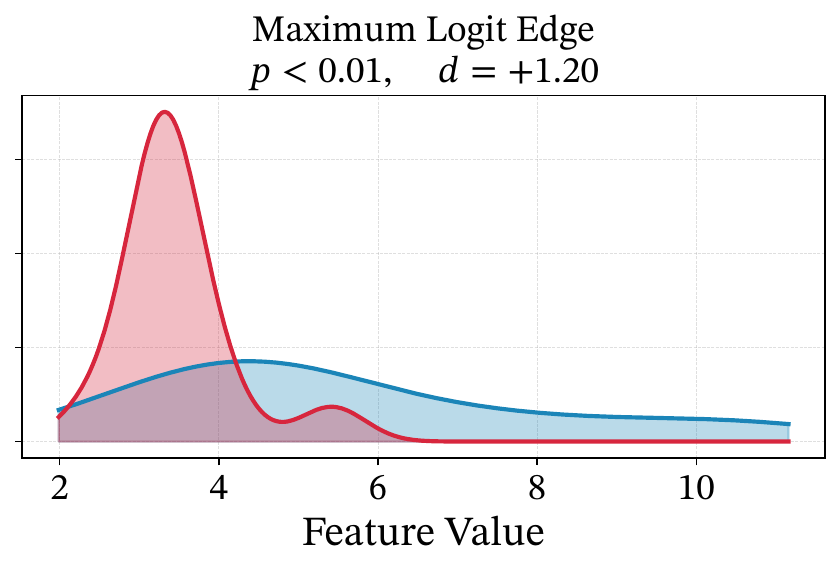}
\end{minipage}\hfill
\begin{minipage}{0.2\textwidth}
    \centering
    \includegraphics[width=\linewidth]{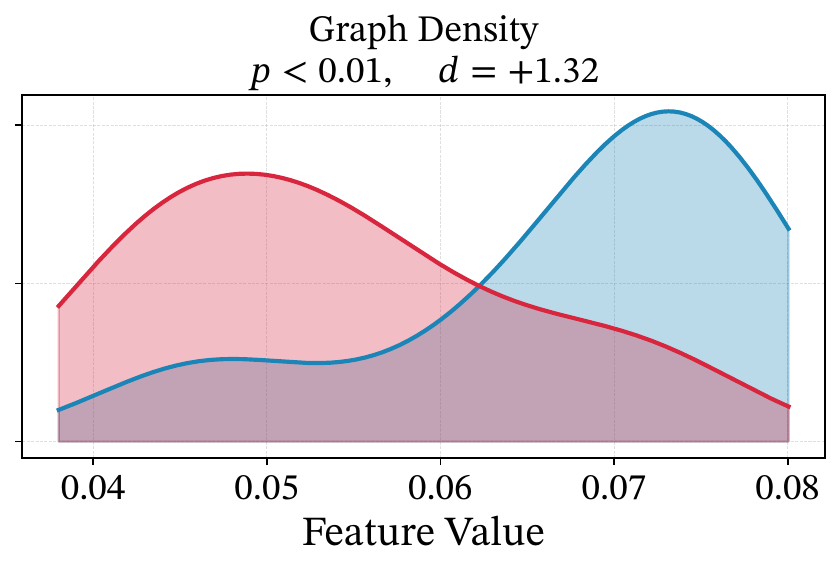}
\end{minipage}
\caption{Distributions of five selected graph-structural features for high- versus low-accuracy probe layers, shown for Toxicity. The separation is statistically significant for each feature ($p < 0.05$, independent $t$-test) and corresponds to medium-to-large effect sizes, with Cohen's $d$ ranging from $+0.88$ to $+1.68$. These results indicate that CTA graphs contain a separable structural signal of a layer's concept-encoding quality.}
\label{fig:topological_fingerprints}
\end{figure*}

We evaluate the linear separability of four concepts across all layers of Gemma-2-2B and Llama-3.2-1B(see Appendix \ref{app:llama}) in order to identify where each concept emerges. Figure~\ref{fig:probe_overview} shows the L0-centred probe score trajectories across layers. Each concept exhibits a distinct belief point: concepts differ sharply in \textit{when} their representations consolidate, not merely in whether they are linearly decodable.

Reasoning shows a distinctive transient suppression: its centred scores fall below zero between L6 and L14 before recovering at L15. This pattern is consistent with multi-hop relational binding requiring deeper computational integration, which temporarily pushes the residual stream toward factual recall. Toxicity and Sentiment consolidate in later layers, whereas Truthfulness reaches its strongest representation in the mid-network~\citep{meng2022neuripslocating}.

\subsection{Structural and Functional Dissociation of Concept and Generation Circuits}
\label{subsection:causal_dissociation}

CTA and LTA graphs extracted from the same prompts show that concept-detection and generation circuits are structurally distinct. Each graph contains a shared computational core, but also recruits a large set of exclusive nodes. Table~\ref{table:causal_dissociation} reports the causal effects of ablating these subspaces.

Ablating probe-exclusive nodes, $G_{\text{probe}} \setminus G_{\text{logit}}$, substantially reduces the concept score across most concepts, confirming that these nodes are causally involved in internal concept encoding. Crucially, this intervention does not affect the generated token. The converse intervention is symmetric: ablating logit-exclusive nodes, $G_{\text{logit}} \setminus G_{\text{probe}}$, flips the generated token in 100\% of cases (92\% in Reasoning) while leaving the concept score essentially unchanged ($\Delta S \approx 0$).

Together, these interventions demonstrate a clean functional double dissociation. The machinery responsible for internally encoding a concept is computationally separable from the machinery responsible for producing the model's output.

The dissociation is weaker for Truthfulness. In this case, probe-exclusive ablation produces a smaller change in concept score, and the shared subgraph carries greater dependence. This is consistent with factual verification requiring more distributed computational resources and drawing on representational machinery that partially overlaps with generation.

The variation in exclusive subgraph size and dissociation strength across concepts suggests that probe-circuit structure reflects the quality and organization of the underlying concept representation. This motivates the next question: whether the topology of the CTA graph, and the specific features exclusive to the probe circuit, can serve as proxies for probe quality at both the layer and prompt levels.

\subsection{Graph-Structural Features Predict Probe Quality}
\label{subsection:graph_structural_features}

\begin{table}[t]
\centering
\small
\begin{tabular}{lcc}
\toprule
\textbf{Model} & \textbf{Spearman} $\rho$ & $R^2$ \\
\midrule
Layer baseline & $0.08 \pm 0.15$ & $-0.19 \pm 0.29$ \\
Ridge          & $0.65 \pm 0.11$ & $0.33 \pm 0.27$ \\
GB             & $\mathbf{0.91} \pm 0.05$ & $\mathbf{0.84} \pm 0.14$ \\
\bottomrule
\end{tabular}
\caption{Predicting probe accuracy from graph-structural features using 5-fold group-aware cross-validation. Bootstrap 95\% confidence intervals for GB are $\rho \in [0.724, 0.964]$, with $p < 0.005$ under group-aware permutation.}
\label{tab:rq2}
\end{table}

We hypothesize that when a model forms a linearly decodable representation of a concept, the computational graph producing that representation should exhibit a recognizable topological signature. This premise parallels recent work on reasoning verification~\citep{zhao2026verifyingchainofthoughtreasoningcomputational}, which shows that valid chain-of-thought processes can be classified largely from their structural graphs. We adapt this idea to internal concept representations: instead of classifying reasoning validity, we ask whether the macroscopic structure of a CTA graph predicts the validation accuracy of the linear probe at that layer.

Graph-structural features predict within-concept layer variation in probe accuracy well beyond a layer-depth baseline. This signal persists under de-meaned targets, as shown in Table~\ref{tab:rq2}. The relationship is robust: the bootstrap lower bound on $\rho$ remains above 0.72, and none of the 200 group-aware permutations approaches the observed value (see Appendix \ref{appendix: detailed results}). This rules out both fold-level artefacts and the trivial explanation that the model is merely exploiting layer depth.

The feature patterns shown in Figure~\ref{fig:topological_fingerprints} suggest that effective concept encoding is characterized by concentrated, decisive circuits. High-accuracy layers exhibit higher Influence Concentration, measured by Gini, indicating that attribution is routed through a smaller number of dominant nodes rather than being distributed diffusely. Maximum Feature Activation is also elevated, showing that concept-relevant features fire more strongly. Maximum Edge Weight and Maximum Logit Edge increase as well, reflecting sharper feature-to-feature routing and a stronger direct read-out connection to the output. Finally, graph density rises as edges accumulate faster than nodes, indicating that the internal connectivity of the graph becomes proportionally tighter at better probe layers. Together, these properties describe circuits that have crystallized around the target concept: sparse but strong, and concentrated rather than broad.

The intensity of this structural signature depends on the concept. For Toxicity, Influence Concentration most clearly separates high- and low-accuracy layers, consistent with a concept whose encoding is dominated by a small number of strongly firing features. For more inferential concepts such as Truthfulness, the signature is weaker across all five features, suggesting that these representations are more broadly distributed and do not collapse into a single concentrated pathway. Thus, the extent to which a concept admits sparse circuit encoding is itself reflected in how clearly structural statistics differentiate strong and weak probe layers. We demonstrate in Appendix \ref{app:llama} that this phenomenon is weaker but persists in Llama-3.2-1B.

\subsection{Feature-Level Prediction of Probe Classification Power}
\label{subsection:feature_level_prediction}

The previous section established that graph-level structural properties characterize strong probe layers. We next ask whether this signal extends to the feature level: can the activation patterns of individual CLT features at a prompt-level predict the probe's classification margin?

\noindent\textbf{Toxicity.}
For Toxicity, a small set of consistent features in the CTA graphs explains a substantial part of the probe score. Two first-layer features carry most of this signal. L0/F8116 has the highest importance and is present in roughly half of the prompts, producing a large shift in probe margin when activated. Verified activating examples show that it fires on derogatory language targeting mental capacity and rationality. L0/F4345, by contrast, activates on strong expletives and targeted slurs. These features show that the toxicity probe margin is already predictable from the first transcoder layer
\begin{figure}[t]
    \centering
    \includegraphics[width=0.9\linewidth]{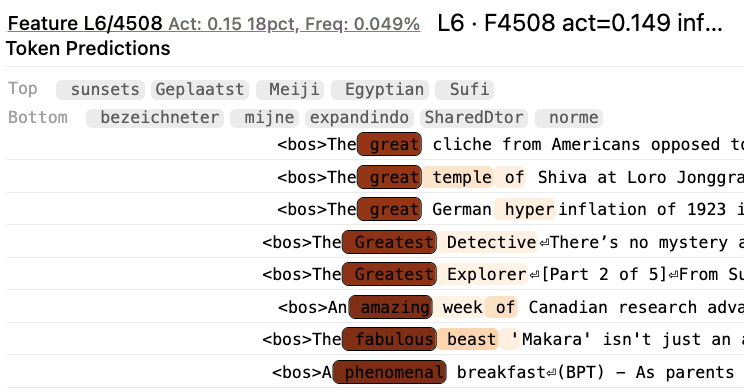}
    \caption{Top activated examples for L6/F4508. This feature is present in CTAs of more than 50\% of positive prompts from the SST-2 sentiment dataset, specifically activating in positive superlatives.}
    \label{fig:feature_sentiment}
\end{figure}
and remains stable across all layers, indicating a concentrated and lexically grounded circuit.

\noindent\textbf{Sentiment.}
For Sentiment, the features most correlated with the probe ($\rho = 0.54$) are primarily localized between L0 and L10. The most informative feature appears in 87 out of 100 cases and fires with near-identical activation on both positive and negative sentiment prompts, suggesting that it detects the domain of the prompt rather than sentiment polarity itself.

Two early-layer features further contribute to the linear direction. L6/F4508 (Figure~\ref{fig:feature_sentiment}) activates on \textit{positive evaluative intensifiers and superlatives}, but also fires on the same words when they occur in factual proper-noun contexts. It therefore partially encodes domain rather than polarity. L7/F5701 captures language associated with emotional and epistemic reactions, including expressions of \textit{surprise}, \textit{regret}, and \textit{suspicion}. This feature skews negative, but does not exclusively mark negative sentiment. We hypothesize that because polarity in film reviews requires sentence-level integration, including metaphor and negation, the attribution-graph feature list only partially recovers the probe direction.

\noindent\textbf{Reasoning.}
The CTA graphs for Reasoning reveal a structural signal that sparse feature presence alone cannot capture. L11/F9661 appears in reasoning attribution graphs across 71 prompts and in zero memorisation graphs (Fisher $p < 0.001$). Verified activating examples show that it fires on binary comparisons, such as \textit{``formal or casual''}, \textit{``fiction from nonfiction''}, and \textit{``whether the system is automated or manual''}. This feature detects the syntactic structure of multi-choice comparative questions, acting as a marker that distinguishes reasoning-requiring prompts from single-hop factual recall. Its signal is real, but it does not generalize into a stable feature-level predictor: the relevant reasoning markers remain prompt-specific.
\begin{figure}[t]
  \centering
  \includegraphics[width=0.8\linewidth]{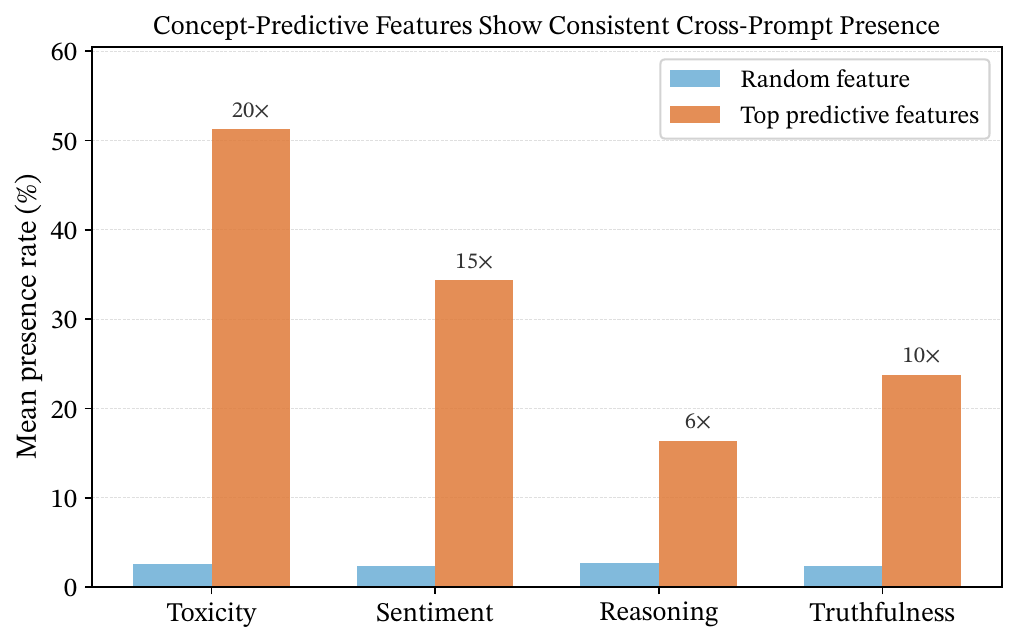}
  \caption{Top predictive features appear 6--20$\times$ more frequently than a random feature at $L_b$, revealing a stable mechanistic core despite prompt-level circuit variation.}
  \label{fig:feature_enrichment}
\end{figure}

\noindent\textbf{Truthfulness.}
The circuit is sparse and mid-layer, with no early binary switch analogous to Toxicity. Its most informative features are falsehood detectors rather than truth detectors. L9/F2578 fires strongly on geographic proper nouns, including city names, country identifiers, and regional markers, making it sensitive to the geographic mismatches that characterize false statements in the dataset, such as \textit{``The Capital of France is Berlin''}. L9/F14198, a truth detector, activates with high precision on newswire-style factual claims, especially institutional facts. L1/F3007 fires on proper names and academic institutional contexts, acting as a dataset-specific truth anchor. Overall, the falsehood detecting features are sharper than the positive ones, suggesting that the circuit is more sensitive to markers of falsehood than to markers of truth.

Across concepts, the few features that appear in multiple circuits specifically mark neutral, analytic, or factual register consistently suppress toxicity while supporting reasoning or truthfulness simultaneously. Features that suppress reasoning while supporting truthfulness suggest that the two concepts recruit distinct lexical cues. We provide some examples and a more detailed view this further at in Appendices \ref{appendix: detailed results} and \ref{app:examples}.

\subsection{Cross-Prompt Consistency of Concept-Attribution Graphs}
\label{subsection:cross_prompt_consistency}

The preceding analyses establish that CTA graphs carry a measurable signal of concept encoding. We now ask how stable these circuits are across prompts. Recent work shows that multiple structurally distinct circuits can perform the same task, challenging the assumption that mechanistic interpretability necessarily recovers a canonical mechanism ~\citep{chen2026circuitsleadromerethinking} . Our results offer a more nuanced picture: while concept-attribution graphs diverge across prompts at the periphery, they converge on a shared feature vocabulary, suggesting a canonical mechanistic core beneath surface variation.

\noindent\textbf{Many Circuits, One Core.}
For each concept, we compute the \textbf{mean presence rate} of the \textit{top predictive features} from our predictor (Section ~\ref{subsection:feature_level_prediction}), by pooling their presence in the CTA graphs of 100 prompts at the belief layer and the layer with the highest validation accuracy. We compare this against the expected presence rate of any other random feature. 

Top predictive features appear 6--21 times more often than a random feature across both layers (Figures~\ref{fig:feature_enrichment},~\ref{fig:feature_enrichment_val}). Toxicity and Sentiment show the strongest and most stable enrichment, driven by a small set of lexically specific features that recur reliably across prompts. Reasoning has lower enrichment, which is expected given how distributed its circuit is. Lastly, Truthfulness shows the largest gap between the two layers, suggesting its predictive features become more prominent deeper in the network.

These results replicate the multi-circuit finding at the level of concept probes: the function of encoding a concept is implemented by many peripherally distinct circuits across prompts, yet they converge on a shared core composed of the features identified as top predictors of the probe margin in Section~\ref{subsection:feature_level_prediction}.

\subsection{Cross-Layer Stability of Concept-Attribution Graphs}
Having established that attribution graphs share a stable feature core across prompts, we measure \textbf{cross-layer stability} via the consecutive-layer Jaccard $J(L, L-1)$, the overlap between attribution graphs at adjacent layers for the same prompt.

\begin{figure*}[t]
\centering
\begin{minipage}{0.25\textwidth}
    \centering
    \includegraphics[width=\linewidth]{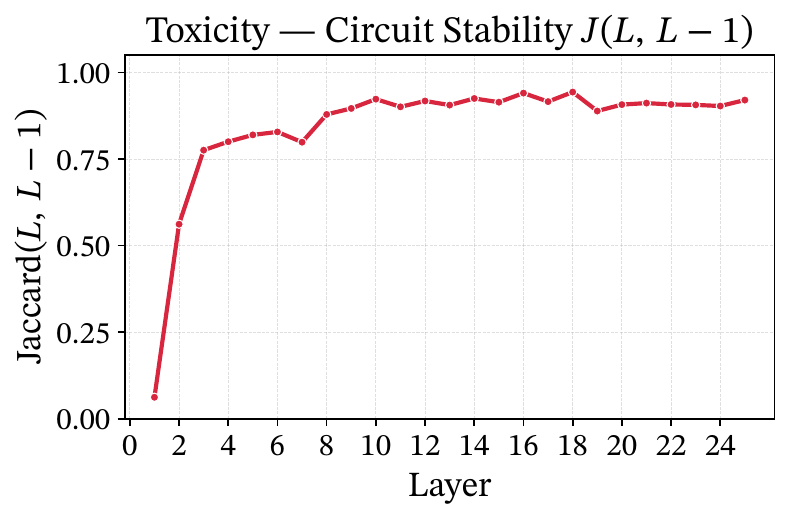}
\end{minipage}\hfill
\begin{minipage}{0.25\textwidth}
    \centering
    \includegraphics[width=\linewidth]{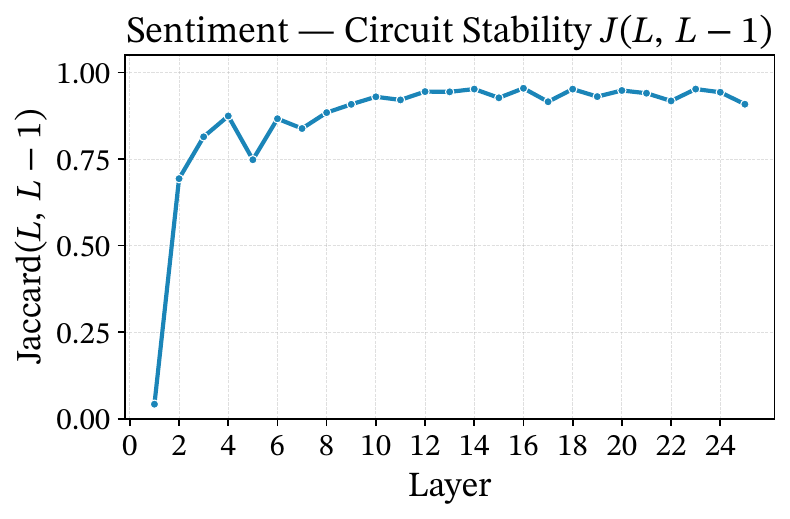}
\end{minipage}\hfill
\begin{minipage}{0.25\textwidth}
    \centering
    \includegraphics[width=\linewidth]{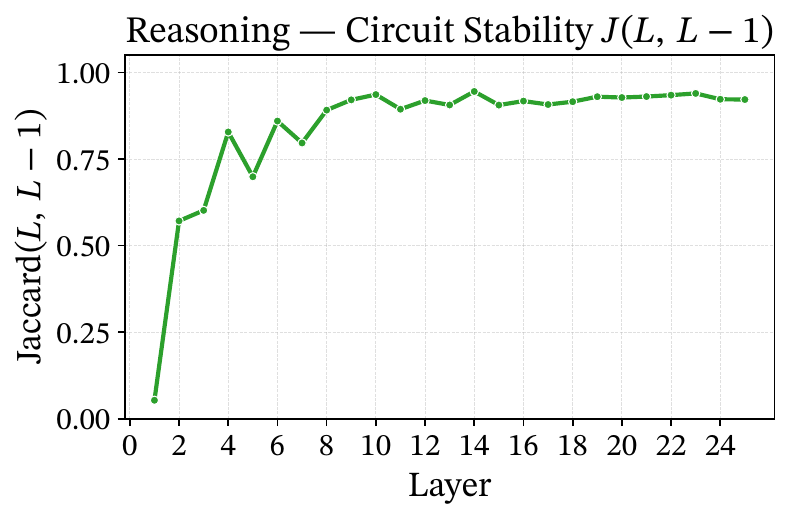}
\end{minipage}\hfill
\begin{minipage}{0.25\textwidth}
    \centering
    \includegraphics[width=\linewidth]{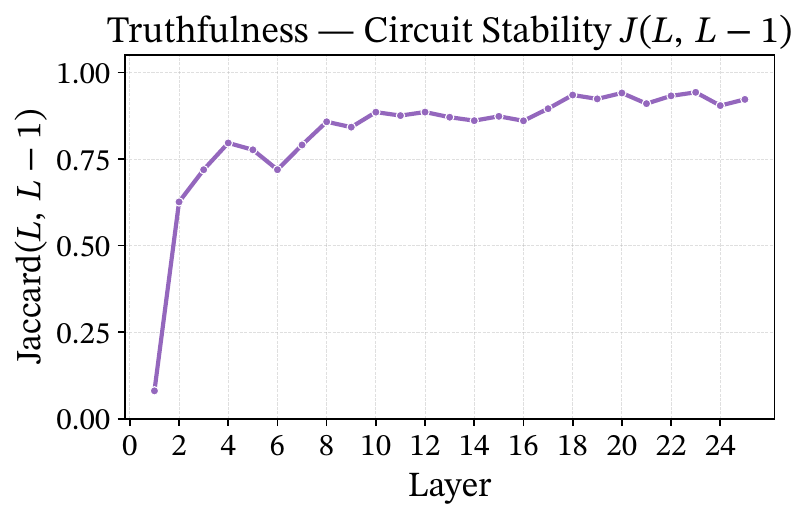}
\end{minipage}\hfill
\caption{Consecutive-layer Jaccard $J(L, L-1)$ for a representative prompt per concept, showing rapid circuit evolution in early layers followed by stabilization at $J \approx 0.9$ by intermediate layers.}
\label{fig:probe_layer_jaccard}
\end{figure*}

As shown in Figure~\ref{fig:probe_layer_jaccard}, $J(L, L-1)$ is low in the early layers, reflecting rapid feature turnover as the circuit is being assembled. The features stabilise at approximately $J=0.9$ between the early intermediate layers(8-12) and remains there through the final layers. This is consistent with our feature-level findings, wherein we observe that the highest-weight predictive features are concentrated at early layers and form the semantic core of the concept circuit, while later layers contribute refinements without displacing the core. 

\section{Discussion}
\label{section:discussion}
\noindent\paragraph{Probes need mechanistic explanations.}
Linear probes have become a standard tool for studying whether neural networks represent concepts, and are increasingly used in safety-relevant settings to monitor properties such as toxicity, truthfulness, or deception. Yet a probe score is a scalar readout, informing the practitioner of \textit{whether} a concept is encoded, not \textit{how} it is encoded. The features causally responsible for the probe score are largely non-overlapping with those that drive generation, a dissociation that holds across Gemma-2-2B and Llama-3.2-1B, suggesting it may be a structural property of the architecture class. CTA provides the mechanistic grounding by identifying the features causally producing the concept representation, enabling targeted steering of belief and generation circuits independently. This opens the possibility of surgical interventions such as suppressing a harmful concept at the representational level without disrupting the model's broader generative behavior.
\noindent\paragraph{Circuit structure as a diagnostic of concept encoding.}
Beyond identifying which features encode a concept, our results demonstrate that the structure of the attribution graph is itself informative. At the feature level, top predictive CLT features are semantically interpretable and concept-specific, appearing 6--21 times more frequently than a random feature across prompts, forming a stable core that the model reliably recruits. At the graph level, structural properties such as mean indegree, and maximum activation predict probe accuracy across layers without access to any labels. Taken together, these two channels offer a dual decomposition: feature identity reveals \textit{what} the model encodes, while graph topology reveals \textit{how well} it has encoded it. Notably, for inferential concepts like Reasoning, where individual features are sparse and distributed, the structural signal generalizes where the feature-level signal weakens. This suggests a practical diagnostic: attribution graph topology can serve as a probe-free indicator of concept internalization at a given layer, with implications for auditing model behavior across training checkpoints or model variants.

\noindent\paragraph{Circuit stability as evidence of structured encoding.}
The cross-prompt and cross-layer consistency results together suggest that concept encoding in transformer-based LLMs is not an opportunistic pattern that varies arbitrarily with input. The shared feature core across prompts, combined with early circuit stabilization that persists through the network, points to a structured, reproducible computation. This suggests that causal claims about concept encoding generalize beyond individual inputs, a necessary condition for mechanistic interpretability to be meaningful in practice.

\section{Conclusion}
\label{section:conclusion}

Understanding how language models internally represent abstract concepts requires more than knowing that a probe fires. We introduce \textbf{Concept-Targeted Attribution }(CTA) to bridge this gap, grounding probe-based findings in causal mechanistic evidence. Across two model families, we find that concept-encoding and generation recruit causally non-overlapping circuits, suggesting that internal belief and output generation are mechanistically separable computations. Concept circuits are neither monolithic nor arbitrary: while attribution graphs vary across prompts, they converge on a shared feature core that strongly drives concept encoding, and whose structural properties alone predict encoding quality without labeled supervision. Together, these findings establish that knowing and saying recruit mechanistically distinct and causally separable circuits, making probe-targeted concept representations a tractable object for causal interpretability and a precise target for intervention.
\section*{Limitations}
\label{section:limitations_ethics}
CTA inherits the quality of the underlying Cross-Layer Transcoder: features that the CLT conflates, omits, or represents polysemantically are invisible to attribution graphs, and the sparsity--reconstruction trade-off shapes which features appear in circuits without concept-specific tuning. Linear probes are additionally a necessary but not sufficient criterion for conceptual validity. For instance, probes trained on correlated label distributions may capture stylistic rather than conceptual features, and our feature-presence analysis discards activation magnitude, missing concepts whose discriminative signal lives in activation geometry rather than feature identity. Future work should pursue broader cross-architecture replication, deeper intervention experiments targeting fine-grained concept steering, and scaling the analysis to richer concept taxonomies beyond the four studied here.

\ifarxiv
% \section*{Author Contributions}\label{sec:contributions}

\section*{Acknowledgment}
This work was supported in part by the German Federal Ministry of Education and Research (BMBF): Tübingen AI Center, FKZ: 01IS18039B; by the Machine Learning Cluster of Excellence, EXC number 2064/1 – Project number 390727645; by Coefficient Giving; by Swiss AI Compute Grants; by Hector Foundation; and by Schmidt Sciences SAFE-AI Grant. Resources used in preparing this research project were provided, in part, by the Province of Ontario, the Government of Canada through CIFAR, and companies sponsoring the Vector Institute.
\fi

\iffalse
\section{End of Main Paper}
\fi

% \bibliographystyle{acl_natbib}
\bibliography{refs_zhijing,refs_causality,refs_cogsci,refs_nlp4sg,refs_semantic_scholar,refs_}

\clearpage
\onecolumn
\addtocontents{toc}{\protect\setcounter{tocdepth}{2}} 
\appendix
\tableofcontents

\newpage
\section{Experimentation Details}
\label{appendix: experimentation details}
\subsection{Dataset Curation}
We evaluate Gemma-2-2B~\citep{gemmateam2024gemma2improvingopen} and Llama-3.2-1B~\citep{grattafiori2024llama3herdmodels}  across four distinct semantic concepts: Toxicity, Sentiment, Reasoning, and Truthfulness. To ensure our probes capture the true conceptual axis rather than spurious dataset artifacts, we train linear classifiers over the residual stream and apply Independent and Identically Distributed (IID) corrections to extract orthogonal direction vectors, $\theta$, for each concept. 

We deliberately select four concepts to span the transformer's computational lifecycle, encompassing factual retrieval to algorithmic synthesis, to late-stage behavioral overrides. We construct the training data to rigorously isolate these specific mechanisms:

\begin{itemize}
    \item \textbf{Reasoning (Algorithmic Logic):} We contrast HotpotQA~\citep{yang-etal-2018-hotpotqa} (positive: multi-hop bridging) against TriviaQA~\citep{joshi-etal-2017-triviaqa} (negative: single-hop factual recall). By anchoring both classes in a general-knowledge QA format, we control for syntactical structure and strictly isolate the computational overhead required for multi-step variable binding. This probe is derived from the aspect of Reasoning-Memorization Interplay~\citep{hong-etal-2025-reasoning}. For example: \textit{The director of Spirited Away also directed which earlier film featuring a young witch?''} (positive) versus \textit{``In which year was the Eiffel Tower completed?''} (negative).
    
    \item \textbf{Truthfulness (Factual Retrieval):} Simple factual probes often overfit to localized, topical subspaces (e.g., geographic knowledge). To force the probe to identify a domain-agnostic factual backbone, we pool and balance three distinct subsets of the Geometry of Truth dataset~\citep{marks2024geometrytruthemergentlinear}: Counterfactuals, Cities, and Spanish-English Translations. For example: \textit{The capital of Japan is Tokyo.''} (true) versus \textit{``The capital of Japan is Osaka.''} (false).
    
    \item \textbf{Sentiment (Affective Baseline):} We use SST-2~\citep{socher-etal-2013-recursive} (positive vs.\ negative movie reviews) to provide a stable, early-layer affective state. This serves as a structural baseline, allowing us to contrast higher-order procedural logic against basic emotional valence. For example: \textit{A beautifully crafted film that lingers long after the credits roll.''} (positive) versus \textit{``A tedious, poorly paced story that squanders every opportunity it is given.''} (negative).
    
    \item \textbf{Toxicity (Behavioral Override):} We use the publicly available Paradetox dataset~\citep{logacheva-etal-2022-paradetox}, which consists of paired toxic and detoxified sentences. Since the semantic topic remains perfectly identical across the pairs, the probe cannot rely on topical keywords; it strictly isolates the latent intent to harm and the corresponding safety filtering mechanism. 
\end{itemize}
\textit{Note: We do not include any specific examples for toxicity in the paper for public safety. Readers may refer to the Paradetox dataset~\citep{logacheva-etal-2022-paradetox} at their discretion. Moreover for probe training and attribution graph generation since we specifically mean to observe the harmful intent of the models, we do not censor these words when we send an input to the LLM.}
\subsection{Probe Training Details}
\label{app:probe_training}

\paragraph{Method.}
All probes are trained as mass-mean (difference-in-means, DIM) linear classifiers~\citep{marks2024geometrytruthemergentlinear,hong-etal-2025-reasoning}, since the DIM direction is more causally implicated in model outputs than logistic-regression probes. For each concept $c$ and layer $\ell$, last-token residual-stream activations are
extracted for a balanced set of positive and negative examples.
The probe direction is
\[
  \theta_{\text{mm}} = \mu^{+} - \mu^{-},
  \quad
  W_c = \theta_{\text{mm}} \,/\, \|\theta_{\text{mm}}\|,
\]
where $\mu^{+}$ and $\mu^{-}$ are the class-conditional means.
$W_c$ is the direction injected into CTA attribution.

\paragraph{IID correction for accuracy evaluation.}
Since $d_\text{model} = 2304$ substantially exceeds the number of training examples,
we apply an LDA correction in a reduced PCA subspace (top 200 components) to obtain
an IID-corrected direction $\theta_{\text{iid}} = \Sigma_w^{-1} \theta_{\text{mm}}$
for accuracy evaluation only.
Classification uses a midpoint threshold
$\tau = (\bar{s}^{+} + \bar{s}^{-}) / 2$
between the positive- and negative-class mean scores.
$W_c$ (raw DIM) is used for CTA attribution throughout; $\theta_{\text{iid}}$ is
not used outside evaluation.

\paragraph{Class balance and train/val split.}
Each concept uses exactly $n$ positive and $n$ negative examples
($n = 5000$ per class, 10000 total), randomly sampled and shuffled with a fixed seed
(seed 42), ensuring no class-imbalance bias in either the probe direction or the
accuracy estimate. Moreover, we filter the data to be $\leq$ 60 tokens, for convenience in interpreting the attribution graphs. An 80/20 train/validation split is applied within each class independently;
validation accuracies for the belief layers are reported in Table~\ref{tab:probe_accuracy}.

\subsection{Cross-Layer-Transcoder}

A cross-layer transcoder (CLT) can be understood as a mechanism for
representing how information carried by MLPs persists, changes, and is reused
across transformer layers.

For a single token position, let
\[
\mathbf{h}_\ell \in \mathbb{R}^{d_{\mathrm{model}}}
\]
denote the input to the MLP at layer $\ell$. Each layer has an encoder that maps
this residual-stream representation into a sparse feature basis:
\[
\mathbf{z}_{\ell}
=
\mathrm{ReLU}\!\left(
\mathbf{W}_{\mathrm{enc}}^{\ell}\mathbf{h}_{\ell}
+
\mathbf{b}_{\mathrm{enc}}^{\ell}
\right)
\in \mathbb{R}^{d_{\mathrm{features}}}.
\]
Here, $\mathbf{z}_\ell$ contains the feature activations used to describe what
the MLP at layer $\ell$ is responding to.

The key difference between a CLT and a layer-local transcoder is that the CLT
does not reconstruct each MLP output using only features from the same layer.
Instead, the reconstructed MLP output at layer $\ell'$ is allowed to depend on
features extracted from all layers up to and including $\ell'$:
\[
\hat{\mathbf{m}}_{\ell'}
=
\sum_{\ell \leq \ell'}
\mathbf{W}_{\mathrm{dec}}^{\ell \rightarrow \ell'}
\mathbf{z}_{\ell}
+
\mathbf{b}_{\mathrm{dec}}^{\ell'} .
\]
The matrix
\[
\mathbf{W}_{\mathrm{dec}}^{\ell \rightarrow \ell'}
\in
\mathbb{R}^{d_{\mathrm{model}} \times d_{\mathrm{features}}}
\]
specifies how features active at layer $\ell$ contribute to the MLP output at
layer $\ell'$.

The fundamental idea is that features are not necessarily confined to a single
layer. A feature may first become active at one layer, then continue to
influence later layers, possibly after being transformed by the model's
intermediate computations. A layer-local transcoder would tend to rediscover
separate versions of this feature at each layer, producing multiple feature
nodes for what is, conceptually, the same underlying computation. A CLT addresses this by allowing earlier-layer features to directly explain
later-layer MLP outputs. This makes it possible to distinguish between genuinely
new features and features that are carried forward from previous layers. As a
result, the attribution graph becomes less redundant: instead of representing
the same semantic signal as several disconnected layer-specific nodes, the CLT
can represent it as a feature whose effects propagate across layers through
learned cross-layer decoder maps \citep{transformer_circuits_jan2025}. We utilise the CLT-Forge library \citep{draye2026clt} for efficient activation loading and open-source CLTs from the Circuit-Tracer library \citep{hanna-etal-2025-circuit}. Training CLTs is computationally expensive, with costs scaling in the number of features and layers. As a result, aside from Anthropic’s foundational work \citep{ameisen2025circuit, lindsey2025biology}, only a small number of CLT applications exist, including a “greater-than” mechanism study \citep{merullo2025replicating}, an open-source extension \citep{lindsey2025landscape}, a multilingual analysis \citep{harrasse2025tracing}, and an attention-computation analysis \citep{draye2025sparse}. This scarcity highlights the difficulty of training CLTs at scale. 
\textit{Note: All datasets, libraries and models are used for research purposes only, consistent with their access conditions.}

\subsection{Attribution Graph}

We construct attribution graphs by assigning a local attribution score between
features at different layers and token positions. Let feature \(n\) at layer
\(\ell\) and position \(k\) be a source node, and let feature \(n'\) at layer
\(\ell'\) and position \(k'\) be a downstream target node. The attribution from
the source feature to the target feature is defined as
\begin{equation}
a^{\ell',k',n'}_{\ell,k,n}
=
\sum_{\ell \leq s \leq \ell'}
\mathbf{f}^{\ell \rightarrow s}_{n}
\,
\mathbf{J}^{\ell',k'}_{s,k}
\,
\mathbf{g}^{\ell'}_{n'} .
\end{equation}
Here, \(\mathbf{f}^{\ell \rightarrow s}_{n}\) is the decoder vector associated
with source feature \(n\) from layer \(\ell\) to layer \(s\), and
\(\mathbf{g}^{\ell'}_{n'}\) is the encoder vector associated with target
feature \(n'\) at layer \(\ell'\).

The term
\[
\mathbf{J}^{\ell',k'}_{s,k}
\]
is the Jacobian of the transformer computation from the MLP output at layer
\(s\), position \(k\), to the MLP input at layer \(\ell'\), position \(k'\).
Importantly, this Jacobian is computed with the model nonlinearities frozen. In practice, the input-dependent components
of the forward pass, including normalization terms, attention computations,
and MLP nonlinearities, are treated as constants using stop-gradient operations.
The resulting Jacobian therefore describes the locally linear propagation of an
MLP-output perturbation through the fixed computation graph.

The sum over \(s\) accounts for the cross-layer structure of the transcoder: a
feature introduced at layer \(\ell\) can contribute to reconstructed MLP outputs
at several later layers \(s\), each of which may influence the target feature at
\((\ell',k')\).

The attribution graph is then obtained by using features at layer-position
pairs as nodes and attribution scores as directed edges. To keep the graph
interpretable, we prune it by retaining only the features that cumulatively
explain \(80\%\) of the effect on the final logit, and the edges that
cumulatively explain \(95\%\) of the retained edge effect on the final logit. This is all done through the Circuit-Tracer library \cite{hanna2025circuit}. 

\subsection{Predictability Experiment Details}
\label{app:predictability}

\subsubsection{Graph Structural Predictor}

We ask whether the topology of a probe attribution graph at layer $\ell$
predicts the probe's validation accuracy at that layer to explain the layer-to-layer variation in probe quality.
We extract 35 graph structural features per prompt (8 global counts, 15 node
statistics, 12 topological measures) and average them across all 100 prompts
at each (concept, layer) pair.
The target is probe validation accuracy; a within-concept demeaned variant
(accuracy minus concept mean) is used as a confound check to prevent the
model exploiting concept-level accuracy offsets.
GradientBoostingRegressor ($200$ estimators, max depth $3$) is evaluated under 5-fold group-aware cross-validation, where groups are (concept,~layer) so that no group seen in training appears in the test fold. A layer baseline model predicting per-layer mean accuracy across concepts is included as a lower bound; GB exceeds it by $\Delta \rho = 0.83$ on the target, confirming the signal is not simply a within-concept layer trend. Statistical validity is assessed via a permutation test (200 group-aware label shuffles, $p < 0.005$) and a bootstrap 95\% confidence interval (500 out-of-bag group resamples: $\rho \in [0.724, 0.964]$ on the target).

\subsubsection{Local CLT Feature Predictor}

As a secondary analysis, we ask whether the identities of CLT features in a
single prompt's attribution graph predict how confidently the probe classifies
that prompt.
The regression target is the probe score margin $y = s - \tau$, where $\tau$
is the midpoint between positive- and negative-class means.
The feature matrix $X \in \{0,1\}^{100 \times 100}$ is a binary presence
indicator over the top-100 activation-ranked CLT features per prompt.
Ridge regression ($\alpha = 10.0$ through a small grid search) is the primary model given $n = p = 100$,
with HistGradientBoosting as a secondary non-linear check; both are evaluated
under 5-fold cross-validation with Spearman~$\rho$ as the primary metric.
The analysis is run independently at every layer $\ell \in \{1, \ldots, 25\}$.

\newpage
\section{Extended Result Discussion}

\begin{figure}[t]
    \centering
    \begin{minipage}{0.32\linewidth}
        \centering
        \includegraphics[width=\linewidth]{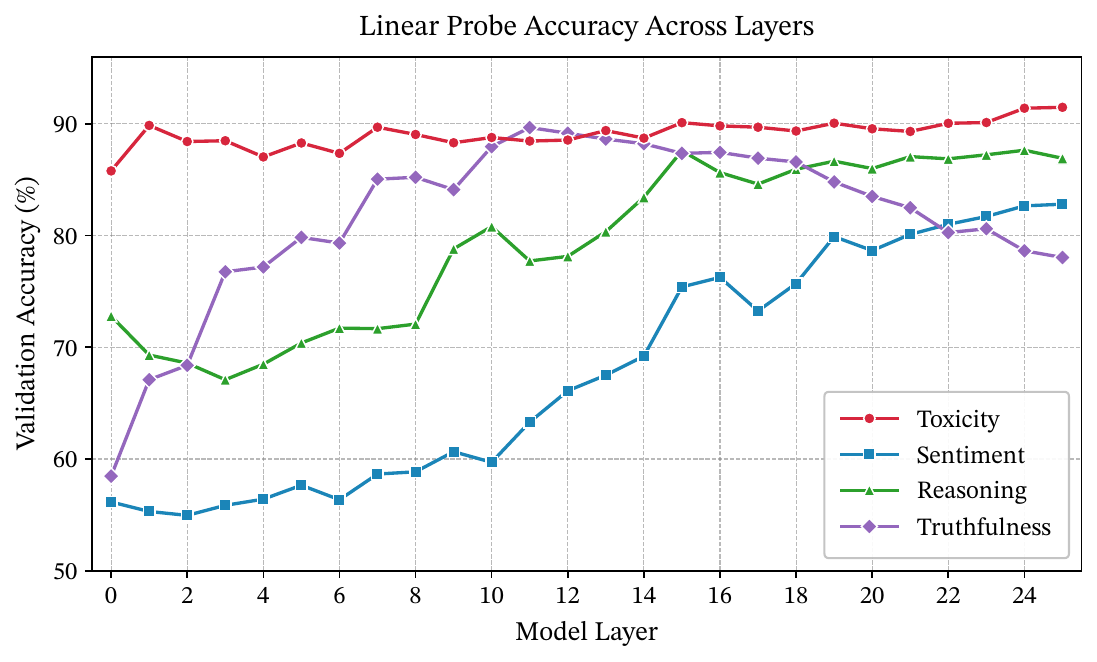}
    \end{minipage}\hfill
    \begin{minipage}{0.32\linewidth}
        \centering
        \includegraphics[width=\linewidth]{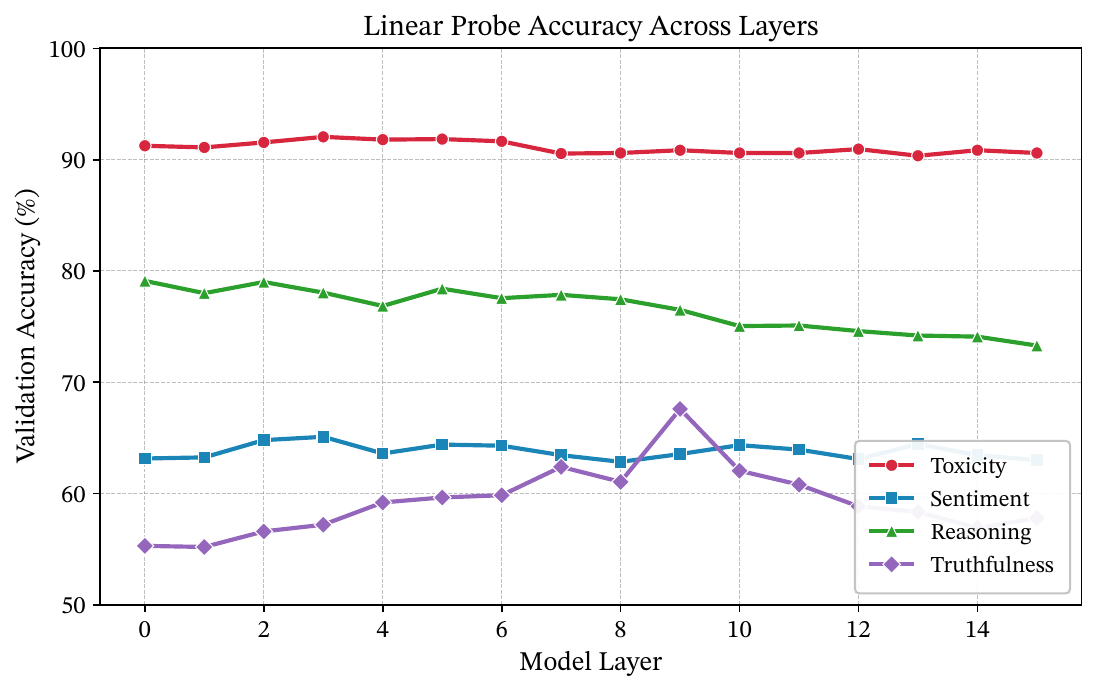}
    \end{minipage}\hfill
    \begin{minipage}{0.32\linewidth}
        \centering
        \includegraphics[width=\linewidth]{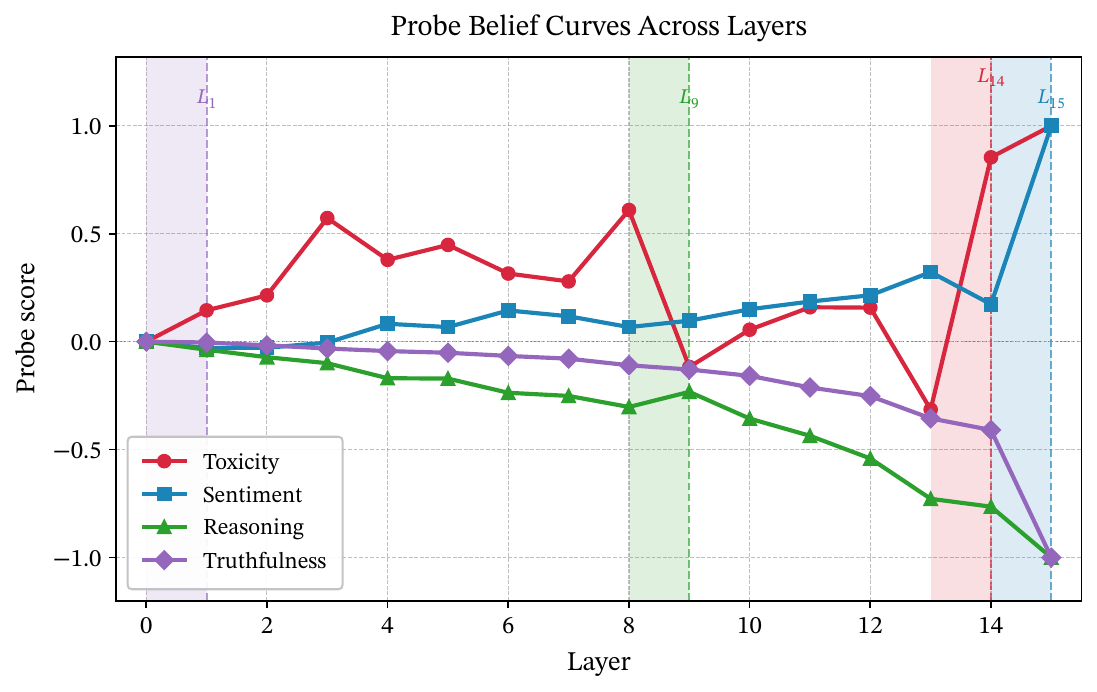}
    \end{minipage}
    \caption{Probe validation accuracy across layers for Gemma-2-2B (left) and Llama-3.2-1B (centre). Toxicity is linearly decodable from early layers in both models; Sentiment and Truthfulness show weaker trajectories in Llama-3.2-1B, consistent with limited concept encoding at the 1B scale. Right: normalised probe score trajectories for Llama-3.2-1B, with shaded bands marking the belief layer $L_b$ per concept.}
    \label{fig:probe_accuracy_trajectories}
\end{figure}

\label{appendix: detailed results}
\subsection{Belief layer selection.}
\label{app:belief}
Probes are trained at every layer $\ell \in \{1, \ldots, 25\}$ and evaluated on the held-out split, but the layer used for CTA attribution is not simply the highest-accuracy layer.
High probe accuracy at a given layer may reflect the residual stream
\emph{retaining} concept information that was encoded many layers earlier, since residual connections accumulate representations cumulatively. A layer can therefore be linearly decodable without being the layer where the concept is actively formed.
We instead identify the \emph{belief layer}: the layer at which the mean
positive-class probe score rises most steeply , which is the point of maximal encoding velocity, where the model is actively constructing the concept direction rather than merely carrying it forward.
In practice, the belief layer is required to also fall within a 2-8\% accuracy plateau of the peak (within-plateau Jaccard $\geq 0.90$ between consecutive layers confirms circuit stability) hence it takes priority over the plateau's highest-accuracy layer.
This ensures that $W_c$ extracted at the belief layer is causally grounded: it captures the direction the model is actively building at that depth, which is the appropriate target for attribution.
Belief layers for Gemma-2-2B are: Toxicity L22, Sentiment L21, Reasoning L15, Truthfulness L14; and for Llama-3.2-1B: Toxicity L14, Sentiment L15, Reasoning L9, Truthfulness L1 (Table~\ref{tab:probe_accuracy}).

\begin{table}[h]
\centering
\normalsize
\begin{tabular}{lcccc}
\toprule
 & \multicolumn{2}{c}{\textbf{Gemma-2-2B}} & \multicolumn{2}{c}{\textbf{Llama-3.2-1B}} \\
\cmidrule(lr){2-3} \cmidrule(lr){4-5}
Concept & $L_b$ & Val acc. & $L_b$ & Val acc. \\
\midrule
Toxicity      & 22 & 90.0\% & 14 & 90.9\% \\
Sentiment     & 21 & 80.1\% & 15 & 63.0\% \\
Reasoning     & 15 & 87.6\% & 9  & 76.5\% \\
Truthfulness  & 14 & 88.2\% & 1  & 55.2\% \\
\bottomrule
\end{tabular}
\caption{Probe accuracy at each concept's belief layer $L_b$ for both models.
Sentiment and Truthfulness show lower ceilings in Llama-3.2-1B, consistent
with weaker encoding of affective and factual concepts at the 1B scale.}
\label{tab:probe_accuracy}
\end{table}

Full ablation results for all four concepts are reported in
Table~\ref{tab:dissociation_full}, extending the study in
Section~\ref{subsection:causal_dissociation}.
Three non-overlapping subsets of each attribution graph are ablated in turn:
probe-exclusive nodes ($G_\text{probe} \setminus G_\text{logit}$),
logit-exclusive nodes ($G_\text{logit} \setminus G_\text{probe}$), and shared
nodes ($G_\text{probe} \cap G_\text{logit}$).
Internal effect is measured as $\Delta S$ (change in probe score at the probe
position); behavioral effect as the top-1 next-token flip rate.

\begin{table*}[htbp]
\centering
\small
\setlength{\tabcolsep}{3pt}
\resizebox{\textwidth}{!}{%
\begin{tabular}{l cccc cccc cccc cccc}
\toprule
& \multicolumn{4}{c}{\textbf{Toxicity} \textit{(L22)}}
& \multicolumn{4}{c}{\textbf{Sentiment} \textit{(L21)}}
& \multicolumn{4}{c}{\textbf{Reasoning} \textit{(L15)}}
& \multicolumn{4}{c}{\textbf{Truthfulness} \textit{(L14)}} \\
\cmidrule(lr){2-5}\cmidrule(lr){6-9}\cmidrule(lr){10-13}\cmidrule(lr){14-17}
\textbf{Subspace}
  & \textbf{N} & $S'$ & $\boldsymbol{\Delta S}$ & \textbf{Flip}
  & \textbf{N} & $S'$ & $\boldsymbol{\Delta S}$ & \textbf{Flip}
  & \textbf{N} & $S'$ & $\boldsymbol{\Delta S}$ & \textbf{Flip}
  & \textbf{N} & $S'$ & $\boldsymbol{\Delta S}$ & \textbf{Flip} \\
\midrule
Control
  & — & $+$102.52 & —              & 0\%
  & — & $+$28.31  & —              & 0\%
  & — & $-$21.44  & —              & 0\%
  & — & $+$41.87  & —              & 0\% \\
$G_{\text{probe}} \setminus G_{\text{logit}}$
  & ${\sim}$190 & $+$41.28  & \textbf{$-$61.24} & 0\%
  & ${\sim}$203 & $-$12.11  & \textbf{$-$40.42} & 0\%
  & ${\sim}$130 & $-$16.39  & \textbf{$+$5.05}  & 0\%
  & ${\sim}$168 & $+$29.84  & \textbf{$-$12.03} & 25\% \\
$G_{\text{logit}} \setminus G_{\text{probe}}$
  & ${\sim}$347 & $+$101.82 & $-$0.70            & \textbf{100\%}
  & ${\sim}$394 & $+$28.70  & $+$0.39            & \textbf{100\%}
  & ${\sim}$539 & $-$20.91  & $+$0.53            & \textbf{92\%}
  & ${\sim}$297 & $+$41.84  & $-$0.03            & \textbf{100\%} \\
$G_{\text{probe}} \cap G_{\text{logit}}$
  & ${\sim}$166 & $+$36.50  & $-$66.02           & 38\%
  & ${\sim}$174 & $+$0.57   & $-$27.74           & 30\%
  & ${\sim}$195 & $-$11.24  & $+$10.20           & 17\%
  & ${\sim}$190 & $+$2.93   & $-$38.94           & 67\% \\
\bottomrule
\end{tabular}%
}
\caption{Full causal ablation results across all four concepts on Gemma-2-2B
  (see Section~\ref{subsection:causal_dissociation} for the main study).
  $S'$ = post-ablation probe score; $\Delta S$ = change from control.
  Reasoning's $\Delta S$ signs are inverted because the L15 probe is
  anti-aligned with all test prompts (baseline $S = -21.44$); ablation
  reduces magnitude rather than score, but the dissociation pattern holds:
  probe-exclusive ablation yields $0\%$ flip, logit-exclusive yields $92\%$
  flip with $\Delta S \approx 0$.
  The logit-exclusive subgraph is $1.77$--$4.15\times$ larger than the
  probe-exclusive subgraph across concepts (Reasoning's $4.15\times$ ratio
  reflects sparse probe encoding at the anti-aligned token position).}
\label{tab:dissociation_full}
\end{table*}

\paragraph{Reasoning: sign inversion and inflated logit ratio.}
Reasoning exhibits two structural differences from the other three concepts.
First, the L15 probe direction is anti-aligned with all test prompts
(baseline probe score $-21.44$; all peak IID values negative), meaning the
probe is reading the \emph{absence} of reasoning signal at the final token
position rather than its presence.
Probe-exclusive ablation consequently raises the score toward zero
($\Delta S = +5.05$), and the sign of all $\Delta S$ entries is inverted
relative to the other concepts — but the qualitative dissociation pattern is
preserved: the probe-exclusive subgraph has zero top-1 flip rate, and the
logit-exclusive subgraph produces a $92\%$ flip rate with near-zero internal
effect ($\Delta S = +0.53$).
Second, the logit-exclusive subgraph (${\sim}539$ nodes) is $4.15\times$ larger
than the probe-exclusive subgraph (${\sim}130$ nodes), substantially above the
$1.77$--$1.94\times$ range observed for the other three concepts.
This inflated ratio reflects that, at this anti-aligned probe position, the
attribution graph assigns more capacity to generation mechanics while probe
representation is sparse which holds true with the concept being weakly
encoded at L15 for these particular prompts rather than a failure of
dissociation itself.

\paragraph{Truthfulness: dominant shared subgraph.}
For Truthfulness, ablating the shared subgraph causes a $93\%$ drop in probe
score ($\Delta S = -38.94$ from a baseline of $+41.87$) with a $67\%$ flip
rate , that is the largest combined effect across both signals of any concept.
The probe-exclusive subgraph produces only a $25\%$ flip rate, unlike the
strict $0\%$ seen for Toxicity and Sentiment, suggesting partial routing of
the truthfulness representation through nodes also used for generation.
The logit-exclusive ablation remains clean ($\Delta S = -0.03$, flip $= 100\%$),
confirming the generation pathway is still separable; the blurring occurs in
the shared component, not in the concept-exclusive one.
\subsection{Graph-Structural Predictability: Gemma-2-2B}
\label{subsection:graph_structural_features_appendix}

Table~\ref{tab:ch2_cv_gemma} reports 5-fold group-aware cross-validation results
for the GradientBoosting regressor on both the raw and de-meaned val\_acc
target. The layer baseline (which predicts the mean val\_acc per layer
across concepts) yields negative R\textsuperscript{2} on the raw target,
ruling out the trivial layer-trend hypothesis. GB substantially exceeds the baseline on both targets.

\begin{table}[t]
\centering
\begin{tabular}{lcccc}
\toprule
Model & \multicolumn{2}{c}{Raw} & \multicolumn{2}{c}{De-meaned} \\
\cmidrule(lr){2-3}\cmidrule(lr){4-5}
 & $\rho$ & $R^2$ & $\rho$ & $R^2$ \\
\midrule
Layer baseline & $+0.083$ & $-0.197$ & $+0.587$ & $+0.240$ \\
Ridge          & $+0.655$ & $+0.325$ & $+0.819$ & $+0.648$ \\
GB             & $+0.905$ & $+0.835$ & $+0.925$ & $+0.850$ \\
\bottomrule
\end{tabular}
\caption{5-fold group-aware CV results for Channel~2 (Gemma-2-2B).}
\label{tab:ch2_cv_gemma}
\end{table}

Table~\ref{tab:ch2_perm_gemma} reports the permutation test and bootstrap
confidence intervals. No permutation out of 200 produced a $\rho$ anywhere
near the observed value; the observed statistic lies $\sim$6 standard
deviations above the null mean.

\begin{table}[t]
\centering
\begin{tabular}{lcc}
\toprule
 & Raw & De-meaned \\
\midrule
Observed $\rho$       & $+0.890$ & $+0.908$ \\
95\% CI on $\rho$     & $[+0.724,\,+0.964]$ & $[+0.817,\,+0.962]$ \\
Null mean $\rho$      & $-0.005 \pm 0.128$  & $-0.001 \pm 0.122$  \\
$p(\rho)$             & $<0.005$            & $<0.005$            \\
\bottomrule
\end{tabular}
\caption{Bootstrap 95\% CI (500 resamples) and permutation test (200 shuffles) for
Channel~2 GB model (Gemma-2-2B).}
\label{tab:ch2_perm_gemma}
\end{table}

\begin{figure*}[htbp]
\centering
\begin{minipage}{0.3\textwidth}
    \centering
    \includegraphics[width=\linewidth]{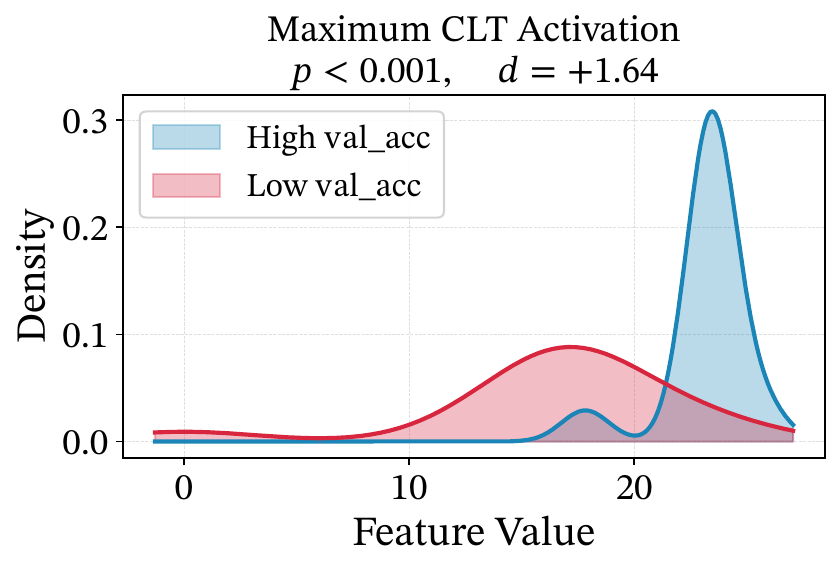}
\end{minipage}\hfill
\begin{minipage}{0.3\textwidth}
    \centering
    \includegraphics[width=\linewidth]{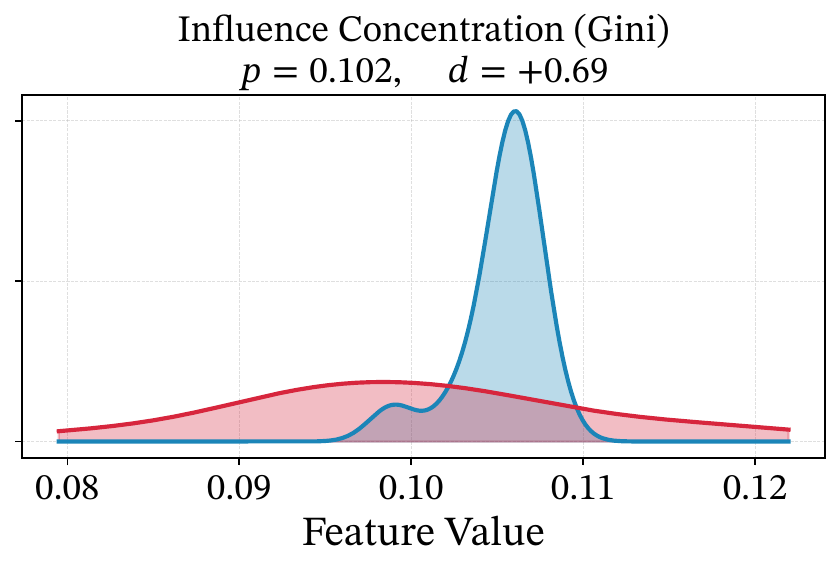}
\end{minipage}\hfill
\begin{minipage}{0.3\textwidth}
    \centering
    \includegraphics[width=\linewidth]{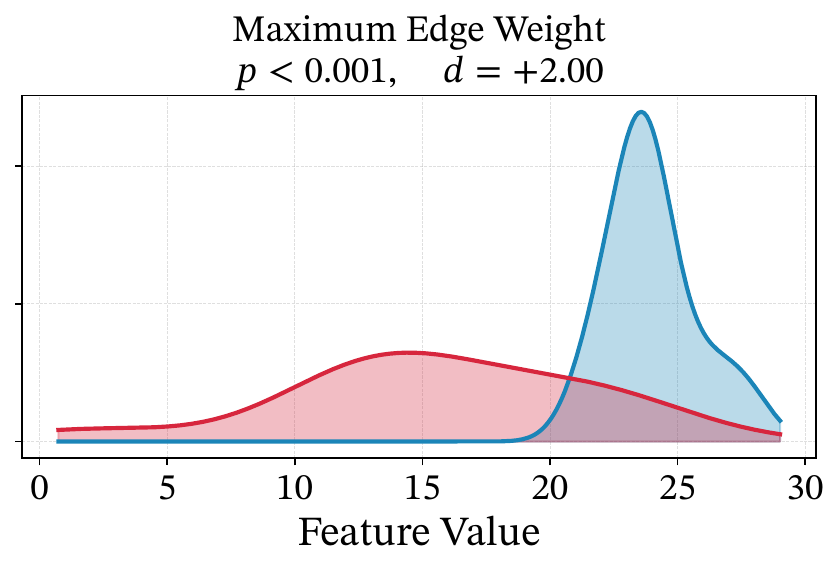}
\end{minipage}\hfill
\begin{minipage}{0.3\textwidth}
    \centering
    \includegraphics[width=\linewidth]{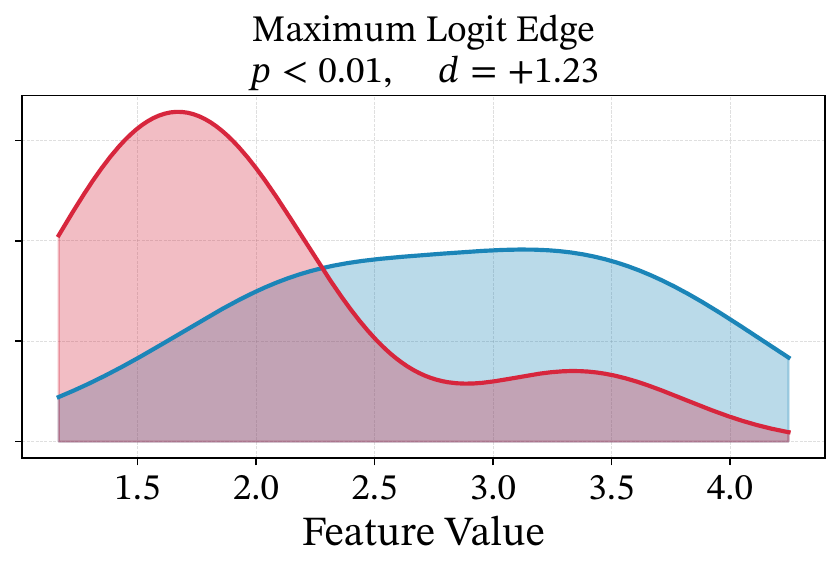}
\end{minipage}
\begin{minipage}{0.3\textwidth}
    \centering
    \includegraphics[width=\linewidth]{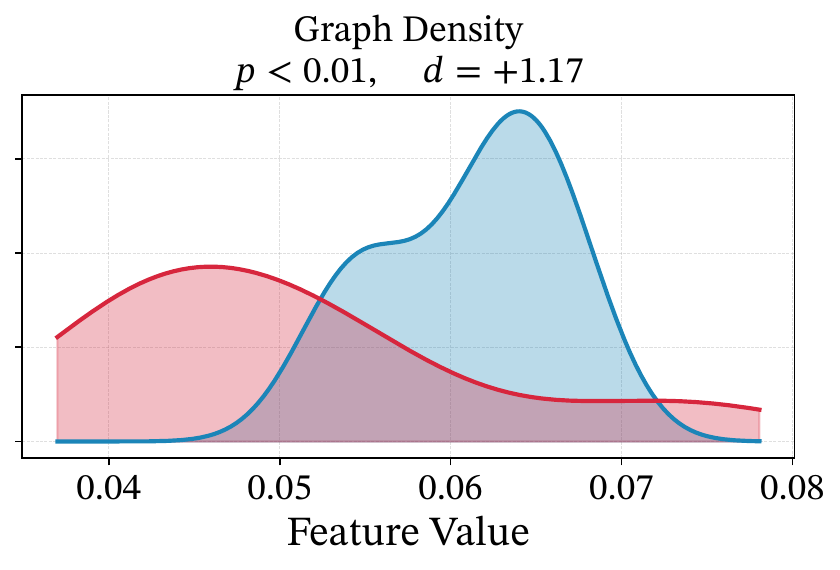}
\end{minipage}
\caption{\textbf{Gemma-2-2B}: Distributions of five selected graph-structural features for high- versus low-accuracy probe layers, shown for reasoning. The separation is statistically significant for each feature ($p < 0.05$, independent $t$-test) except the influence concentration and corresponds to medium to large effect sizes, with Cohen's $d$ ranging from $+0.69$ to $+2.00$. These results indicate that CTA graphs contain a separable structural signal of a layer's concept-encoding quality.}
\label{fig:topological_fingerprints_reasoning}
\end{figure*}

\begin{figure*}[htbp]
\centering
\begin{minipage}{0.3\textwidth}
    \centering
    \includegraphics[width=\linewidth]{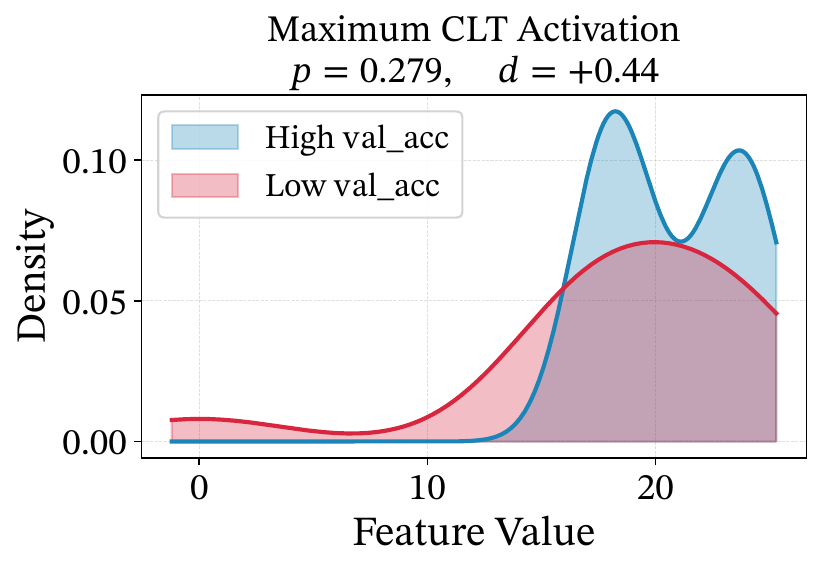}
\end{minipage}\hfill
\begin{minipage}{0.3\textwidth}
    \centering
    \includegraphics[width=\linewidth]{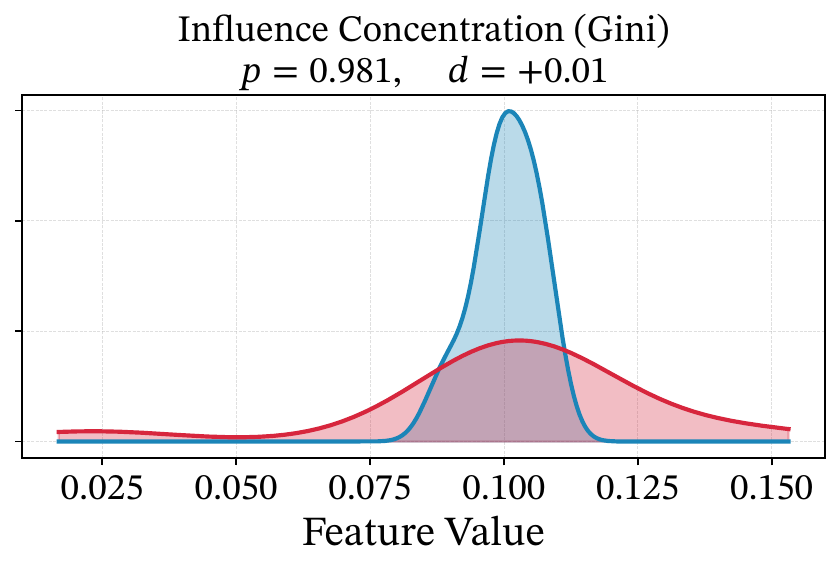}
\end{minipage}\hfill
\begin{minipage}{0.3\textwidth}
    \centering
    \includegraphics[width=\linewidth]{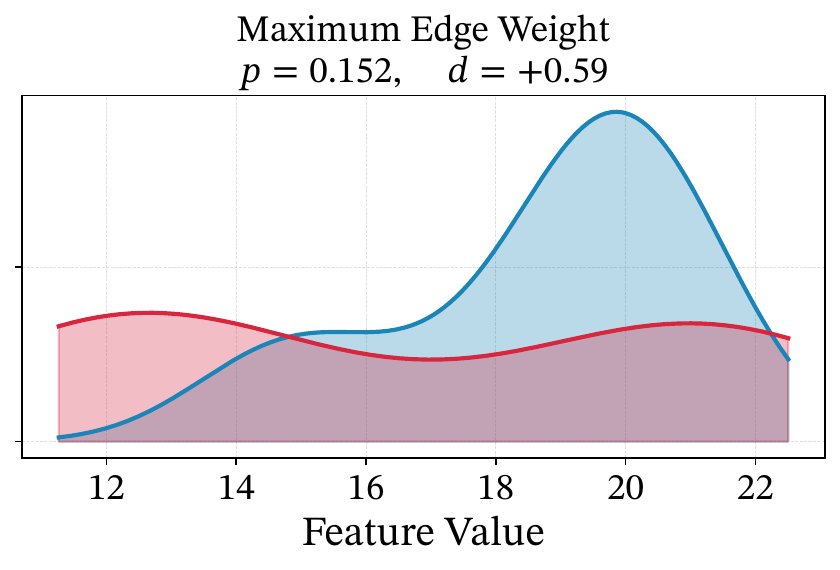}
\end{minipage}\hfill
\begin{minipage}{0.3\textwidth}
    \centering
    \includegraphics[width=\linewidth]{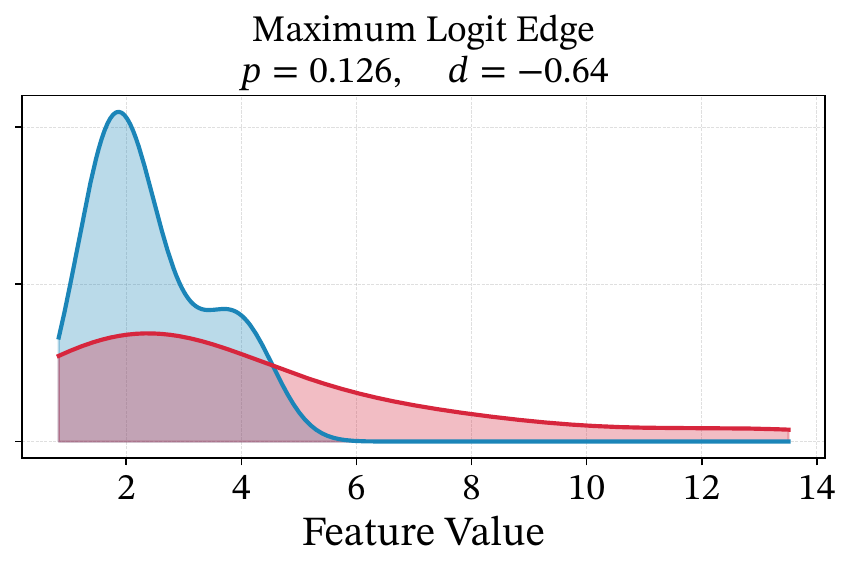}
\end{minipage}
\begin{minipage}{0.3\textwidth}
    \centering
    \includegraphics[width=\linewidth]{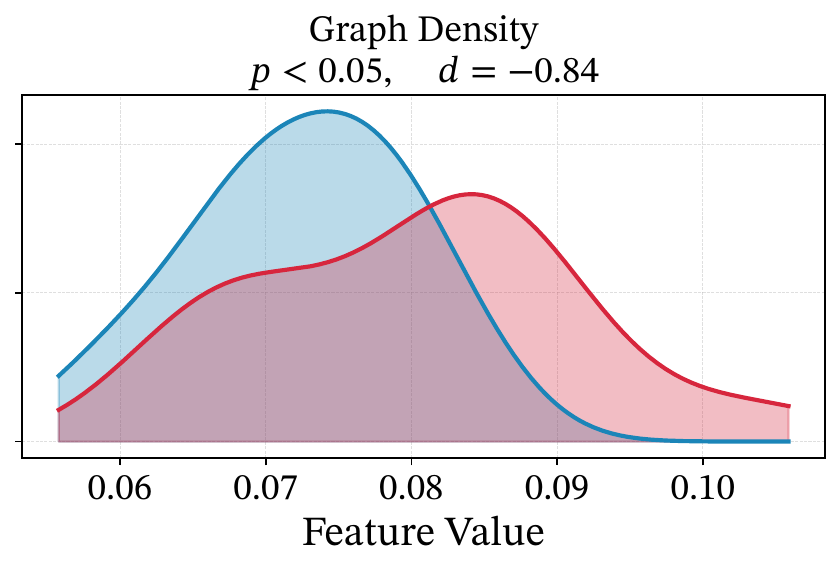}
\end{minipage}
\caption{\textbf{Gemma-2-2B}: Distributions of five selected graph-structural features for high- versus low-accuracy probe layers, shown for truthfulness. The separation is not purely significant for features ($p < 0.05$, independent $t$-test) except graph density and corresponds to low to medium effect sizes, with Cohen's $d$ ranging from $-0.84$ to $+0.59$. These results indicate that CTA graphs contain a separable structural signal of a layer's concept-encoding quality.}
\label{fig:topological_fingerprints_truthfulness}
\end{figure*}

The topological fingerprint distributions for Reasoning, Sentiment, and
Truthfulness are shown in
Figures~\ref{fig:topological_fingerprints_reasoning},
\ref{fig:topological_fingerprints_sentiment},
and~\ref{fig:topological_fingerprints_truthfulness} respectively.

\subsection{Detailed Feature-Level Study}
\paragraph{Toxicity} achieves the strongest and most consistent predictability, with Spearman $\rho$ above 0.66 at every layer and peaking at 0.86 at L10. The signal is dominated by two features encoded at the earliest transcoder layer: L0/F8116 and L0/F4345, each present in roughly half the prompts. When either fires, the predicted probe margin shifts by over 17 points and flips sign in their absence, a near-binary discrimination. Verified activating examples show these features encode semantically specific derogatory language: L0/F8116 targets slurs aimed at mental capacity, and L0/F4345 encodes explicit expletives and targeted slurs. A mid-layer feature L5/F6028 reinforces the signal in prompts involving contextual harm framing, while L14/F13558 contributes at the probe layer itself. This discriminative power holds across all 25 layers, confirming that the toxicity probe's confidence is largely explained by a small set of early-layer lexical features stable throughout the residual stream.
\begin{figure}[t]
  \centering
  \includegraphics[width=0.5\linewidth]{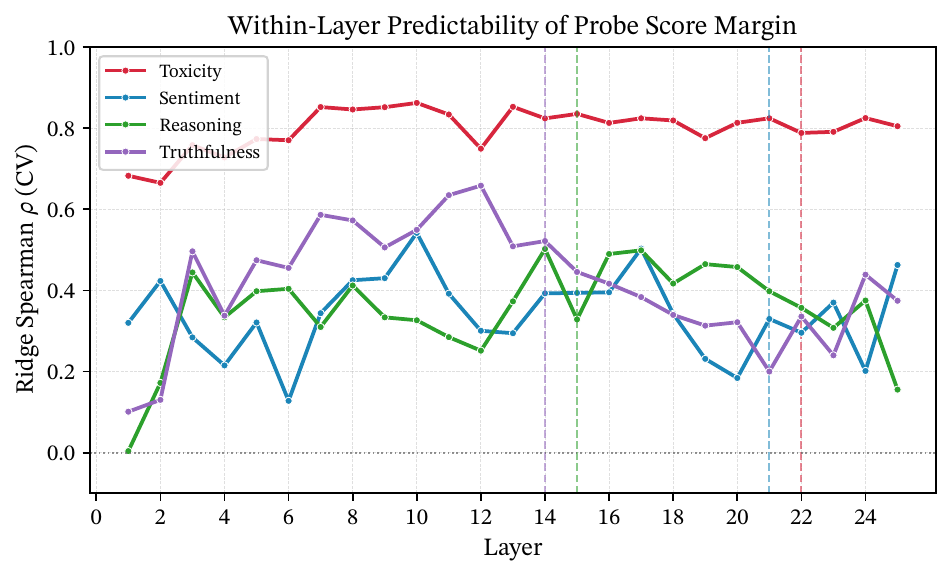}
  \caption{Spearman $\rho$ between predicted and actual probe score margins across layers for all four concepts. Toxicity maintains consistently high predictability across all layers; Truthfulness peaks at mid-depth before decaying; Sentiment and Reasoning show weak and noisy signals throughout, reflecting distributed or geometrically encoded concept representations that individual CLT features cannot recover.}
  \label{fig:tr_heatmap}
\end{figure}
\paragraph{Sentiment} presents the sharpest contrast. Ridge $R^2$ peaks modestly at L10 (0.23) before collapsing to 0.08 at the probe's best layer (L21), with scattered negative $R^2$ at several intermediate layers. The highest-coefficient feature L0/F4345, shared with the Toxicity circuit, is present in 87 of 100 prompts and fires with near-identical activation on both positive and negative reviews, encoding the evaluative film-review register rather than sentiment direction. A mid-layer feature L6/F4508 capturing evaluative intensifiers also fires in factual proper-noun contexts; L7/F5701 encodes epistemic reaction language that skews toward negative affect without exclusively marking it. Polarity in this domain requires integrating sentence-level structure including metaphor, hedging, and clause-level negation, all of which are invisible to individual feature presence. The probe reads a continuous geometric direction in the full residual stream that the discrete feature list cannot recover.

\paragraph{Reasoning} shows the weakest Ridge $R^2$ throughout (peak 0.14), but the complementary rise feature analysis reveals a structural signal. Features at mid-to-late layers (L14/F11925, L10/F6578, L12/F15646) are strongly enriched in prompts where the probe transitions from incorrect to correct, with odds ratios up to 8.8$\times$ relative to consistently low-performing prompts. An early-layer feature L0/F3969, consistently present across reasoning attribution graphs but absent from memorisation graphs, fires specifically on binary comparison constructions such as \textit{formal or casual} and \textit{fiction from nonfiction}, encoding the syntactic architecture of multi-choice questions rather than their content. L6/F7507 contributes a late-activating signal tied to step-by-step elaboration structure. This signal is real but too sparse to elevate $R^2$.

\paragraph{Truthfulness} occupies an intermediate position: Spearman $\rho$ peaks at 0.66 at L12, with meaningful Ridge signal from L7 onward before decaying past L15. The signal is spread across sparse features at mid-range layers; L9/F2578, L9/F14198, and L9/F15693 each fire in only 14--21 of 100 prompts but produce sharp margin shifts when active. L1/F3007 is an exception, present in 49 of 100 prompts, and carries one of the strongest positive coefficients, encoding attributed factual claims and neutral encyclopedic register. The negative-coefficient features consistently outweigh positive-class predictors in magnitude, suggesting the circuit is more sensitive to markers of falsehood than of truth.

\paragraph{Cross-concept features.} A small number of features appear across multiple concept circuits with semantically interpretable cross-concept signs. L1/F3872 carries a negative coefficient in the Toxicity regression and a positive coefficient in the Reasoning regression: a feature firing on well-formed analytic language plausibly suppresses toxicity while supporting reasoning-positive prompts. L0/F16252 and L5/F9293 are both Toxicity-negative and Truthfulness-positive, consistent with a feature encoding neutral, factual statement language that marks non-toxic and truthful text simultaneously. L5/F11818 is Sentiment-negative and Reasoning-positive, suggesting a feature firing on analytical or critical language that is both unpleasant-sounding and reasoning-coded. The structurally interesting case is L9/F15693: it suppresses reasoning-positive probability while supporting truthfulness-positive probability. Since reasoning and truthfulness might be expected to positively correlate, as both favour well-structured and accurate text, their anti-correlation at this feature suggests that the lexical cues for attributed factual claims and for step-by-step inferential structure are partially dissociated at the feature level.

\subsection{Cross-Prompt Consistency of CTA Graphs}
\begin{figure}[t]
  \centering
  \includegraphics[width=0.7\linewidth]{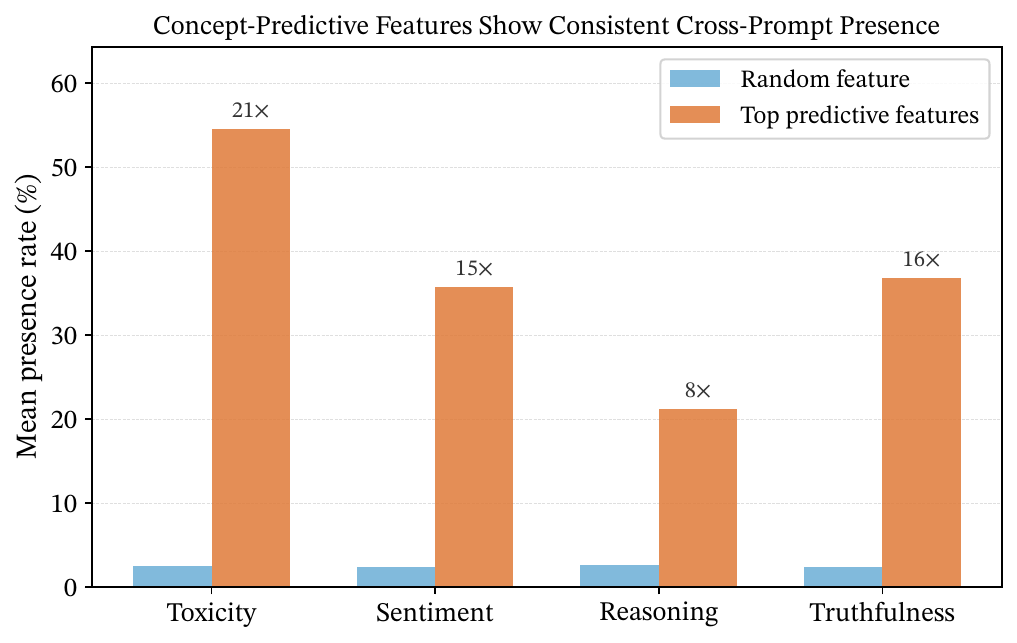}
  \caption{Top predictive features appear 8--21$\times$ more frequently than a random feature at highest validation accuracy layer, revealing a stable mechanistic core despite prompt-level circuit variation.}
  \label{fig:feature_enrichment_val}
\end{figure}

Extending from Section \ref{subsection:cross_prompt_consistency}, we provide the mean presence rate of top predictive features compared with expected rates of random features at the layer with the highest validation accuracy all concepts. The figure is provided at Figure \ref{fig:feature_enrichment_val}.

\subsection{Case Study: Truthfulness vs. Reasoning}
\label{subsubsection:truthfulness_reasoning_case_study_appendix}

Truthfulness and Reasoning are the semantically closest concept pair in our study. Both concern the epistemic quality of language, and their belief layers are adjacent: L14 for Truthfulness and L15 for Reasoning. If any two concepts were likely to share circuit vocabulary, these would be the strongest candidates. We therefore examine cross-concept circuit overlap directly. For every prompt, we compute the per-prompt Jaccard similarity $J(T{@}L_T,\, R{@}L_R)$ across all $(L_T, L_R)$ layer-pair combinations (Figure~\ref{fig:tr_heatmap}).
\begin{figure}[t]
  \centering
  \includegraphics[width=0.5\linewidth]{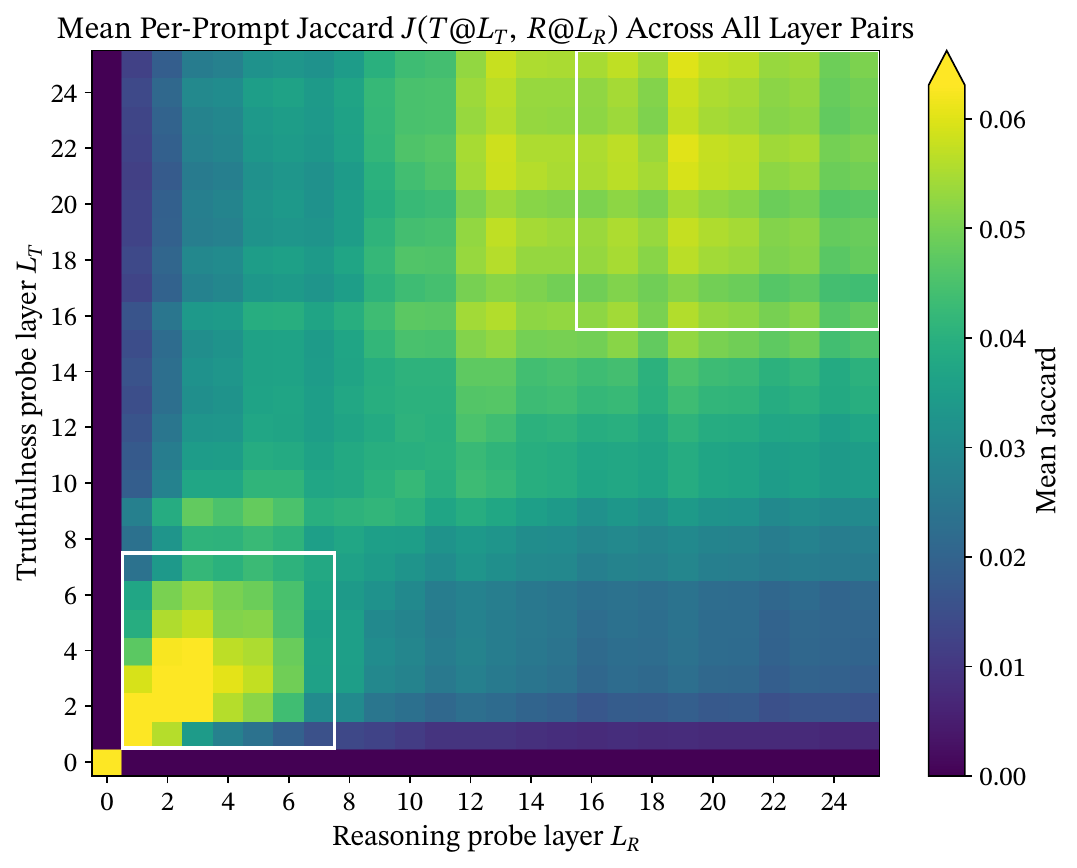}
  \caption{Mean per-prompt Jaccard similarity $J(T{@}L_T,\, R{@}L_R)$ between Truthfulness and Reasoning attribution graphs across all layer-pair combinations. Cross-concept overlap is non-negligible only when both layers are in the same depth regime.}
  \label{fig:tr_heatmap}
\end{figure}
\paragraph{Layer alignment is necessary but not sufficient.}
The heatmap in Figure~\ref{fig:tr_heatmap} reveals two regions of elevated cross-concept Jaccard: early layers ($L_T, L_R \lesssim 7$) and late layers ($L_T, L_R \gtrsim 16$). Off-diagonal entries, such as early Truthfulness against late Reasoning or late Truthfulness against early Reasoning, collapse toward zero. Layer alignment is therefore a prerequisite for cross-concept sharing, reflecting that features at different depths encode qualitatively different representations. However, even on the diagonal, peak values remain below 0.07, well within the low-overlap regime observed in the within-concept analysis.

\paragraph{Exclusive features dominate at every layer.}
At any given layer, the mean number of shared features per prompt remains below approximately 40, whereas reasoning-exclusive counts reach roughly 460 and truthfulness-exclusive counts reach roughly 250 by the final layers. Shared features therefore represent only a small fraction of each concept's circuit throughout the network. The early-layer sharing visible in the heatmap corresponds to a modest absolute count, likely reflecting a small set of syntactic or domain-general features active in both attribution graphs. The bulk of each circuit remains exclusive.

Together, these results extend the many-circuits finding across concept boundaries. Even the most semantically proximate concept pair recruits largely disjoint attribution graphs, with overlap depending on layer alignment and never exceeding a small minority of each circuit's features.

\newpage
\section{Activated Examples for Selected Features}
\label{app:examples}
\begin{figure*}[ht]
\centering
\begin{minipage}{0.4\textwidth}
    \centering
    \includegraphics[width=\linewidth]{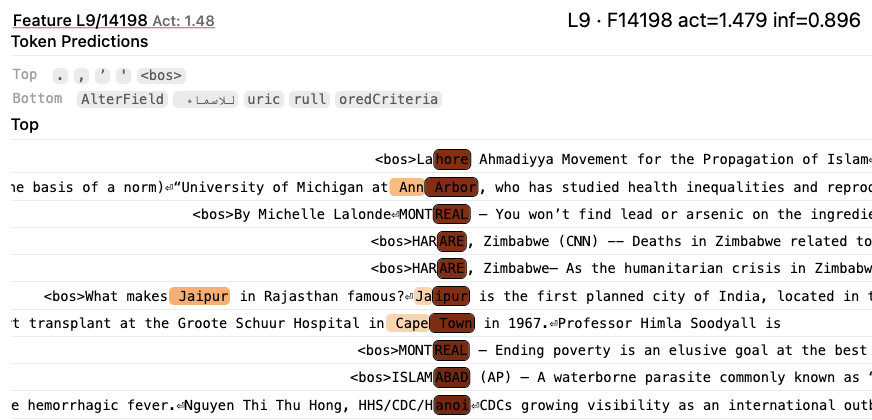}
\end{minipage}\hfill
\begin{minipage}{0.4\textwidth}
    \centering
    \includegraphics[width=\linewidth]{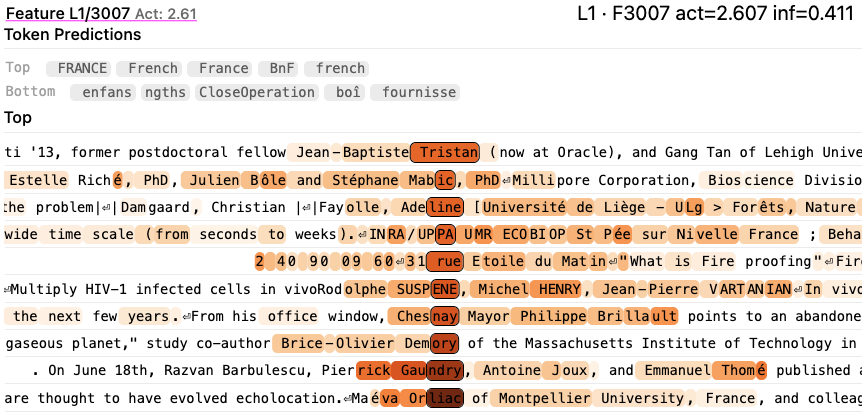}
\end{minipage}\hfill
\begin{minipage}{0.4\textwidth}
    \centering
    \includegraphics[width=\linewidth]{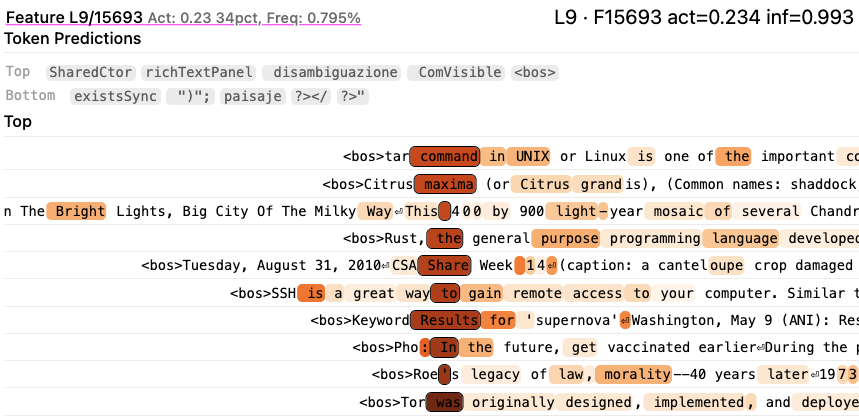}
\end{minipage}\hfill
\begin{minipage}{0.4\textwidth}
    \centering
    \includegraphics[width=\linewidth]{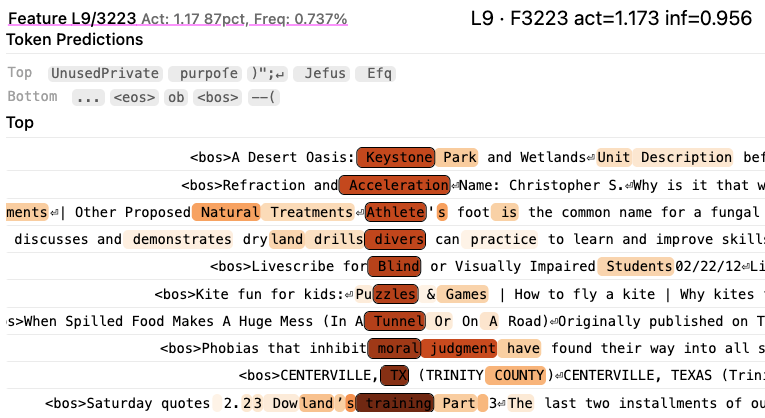}
\end{minipage}
\begin{minipage}{0.4\textwidth}
    \centering
    \includegraphics[width=\linewidth]{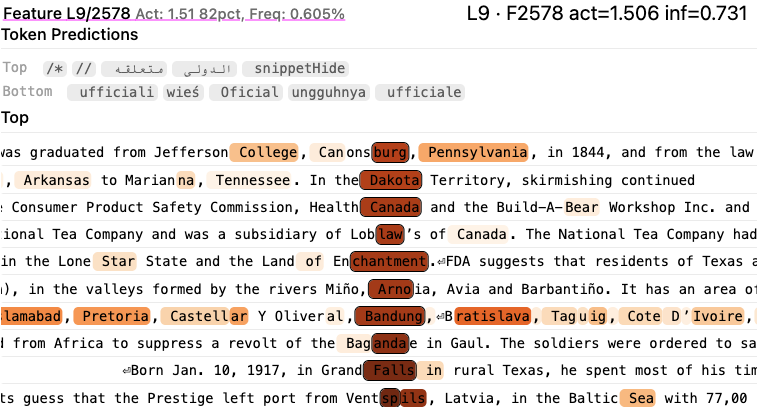}
\end{minipage}
\caption{Representative CLT features identified by the probe attribution pipeline, visualized via Neuronpedia. Each panel shows the feature's layer, index, mean activation, and top/bottom predicted tokens, with activating contexts highlighted. The features encode semantically coherent concepts: L9/F3223 fires on content keywords in structured titles; L9/F2578 and L9/F14198 specialize in geographic place names and city tokens respectively; L1/F3007 captures French proper names and French-language text; and L9/F15693 activates on common English function words across diverse domains. The semantic breadth of these features illustrates that probe-predictive circuits draw on general linguistic primitives alongside concept-specific representations.}
\label{fig:topological_fingerprints_sentiment}
\end{figure*}
\newpage
\section{Implementation Results on Llama-3.2-1B}
\label{app:llama}

\subsection{Causal Dissociation}
\label{app:llama_dissociation}

Table~\ref{tab:dissociation_llama} reports the full causal ablation results
for Llama-3.2-1B across all four concepts. The core dissociation pattern
replicates cleanly: ablating probe-exclusive features ($G_{\text{probe}}
\setminus G_{\text{logit}}$) produces a measurable internal effect on the
probe score while causing $0\%$ top-1 prediction flips across all concepts;
ablating logit-exclusive features ($G_{\text{logit}} \setminus G_{\text{probe}}$)
causes $100\%$ flips while leaving the probe score near-baseline.

Probe IID magnitudes in Llama-3.2-1B are $10$--$20\times$ smaller than in
Gemma-2-2B; thresholds are scaled accordingly and all prompts pass the
strong IID threshold.
The sentiment probe encodes positive sentiment, so all negative-sentiment
prompts yield a negative baseline score; ablating probe-exclusive features
causes the score to rise toward zero, which is the correct direction.
Truthfulness at $L_1$ has only $\sim$11 probe-exclusive nodes and a baseline
accuracy of $55.2\%$. The probe circuit is too weak at this layer to show
a measurable internal effect, though the logit dissociation ($100\%$ flip)
holds cleanly. The logit-exclusive subgraph is $0.83$--$3.23\times$ the
size of the probe-exclusive subgraph (excluding the Truthfulness outlier
driven by the near-zero probe circuit at $L_1$; mean ratio $1.93\times$),
consistent with the Gemma-2-2B pattern.

\begin{table*}[t]
\centering
\small
\setlength{\tabcolsep}{3pt}
\resizebox{\textwidth}{!}{%
\begin{tabular}{l cccc cccc cccc cccc}
\toprule
& \multicolumn{4}{c}{\textbf{Toxicity} \textit{(L14)}}
& \multicolumn{4}{c}{\textbf{Sentiment} \textit{(L15)}}
& \multicolumn{4}{c}{\textbf{Reasoning} \textit{(L9)}}
& \multicolumn{4}{c}{\textbf{Truthfulness} \textit{(L1)}} \\
\cmidrule(lr){2-5}\cmidrule(lr){6-9}\cmidrule(lr){10-13}\cmidrule(lr){14-17}
\textbf{Subspace}
  & \textbf{N} & $S'$ & $\boldsymbol{\Delta S}$ & \textbf{Flip}
  & \textbf{N} & $S'$ & $\boldsymbol{\Delta S}$ & \textbf{Flip}
  & \textbf{N} & $S'$ & $\boldsymbol{\Delta S}$ & \textbf{Flip}
  & \textbf{N} & $S'$ & $\boldsymbol{\Delta S}$ & \textbf{Flip} \\
\midrule
Control
  & — & $+0.43$ & —             & 0\%
  & — & $-0.47$ & —             & 0\%
  & — & $+0.60$ & —             & 0\%
  & — & $+0.11$ & —             & 0\% \\
$G_{\text{probe}} \setminus G_{\text{logit}}$
  & ${\sim}$244 & $-0.10$ & $\mathbf{-0.53}$  & 0\%
  & ${\sim}$168 & $+2.40$ & $\mathbf{+2.87}$  & 0\%
  & ${\sim}$115 & $+0.39$ & $\mathbf{-0.21}$  & 0\%
  & ${\sim}$11  & $+0.11$ & $\mathbf{-0.001}$ & 0\% \\
$G_{\text{logit}} \setminus G_{\text{probe}}$
  & ${\sim}$203 & $+0.43$ & $+0.002$           & \textbf{100\%}
  & ${\sim}$289 & $-0.48$ & $-0.011$           & \textbf{100\%}
  & ${\sim}$372 & $+0.57$ & $-0.028$           & \textbf{100\%}
  & ${\sim}$359 & $+0.09$ & $-0.020$           & \textbf{100\%} \\
$G_{\text{probe}} \cap G_{\text{logit}}$
  & ${\sim}$169 & $+0.67$ & $+0.24$            & 78\%
  & ${\sim}$112 & $+0.45$ & $+0.92$            & 25\%
  & ${\sim}$206 & $-0.33$ & $-0.93$            & 58\%
  & ${\sim}$12  & $+0.08$ & $-0.026$           & 17\% \\
\bottomrule
\end{tabular}%
}
\caption{Full causal ablation results across all four concepts on Llama-3.2-1B.
  $S'$ = post-ablation probe score; $\Delta S$ = change from control.
  Sentiment baseline is negative as the probe encodes positive sentiment and
  all prompts are negative-sentiment text; ablating probe-exclusive features
  correctly reduces the magnitude. Truthfulness at $L_1$ shows a near-zero
  internal effect due to the weak probe circuit ($\sim$11 nodes, $55.2\%$
  accuracy) at that layer; the logit dissociation holds cleanly ($100\%$ flip).}
\label{tab:dissociation_llama}
\end{table*}

\subsection{Graph-Structural Predictability: Llama-3.2-1B}
\label{app:llama_ch2}

Table~\ref{tab:ch2_cv_llama} reports 5-fold group-aware CV results for the
GradientBoosting regressor on the raw \texttt{val\_acc} target.
The layer baseline yields negative R\textsuperscript{2} ($-0.66$), ruling out
the trivial layer-trend hypothesis. GB achieves $\rho = +0.808$ and
$R^2 = +0.850$, replicating the direction of the Gemma-2-2B Channel~2 finding.

\begin{table}[h]
\centering\small
\begin{tabular}{lcc}
\toprule
Model & $\rho$ & $R^2$ \\
\midrule
Layer baseline & $-0.437$ & $-0.660$ \\
Ridge          & $+0.423$ & $+0.069$ \\
GB             & $+0.808$ & $+0.850$ \\
\bottomrule
\end{tabular}
\caption{5-fold group-aware CV results for Channel~2 (Llama-3.2-1B).}
\label{tab:ch2_cv_llama}
\end{table}

\begin{table}[h]
\centering\small
\begin{tabular}{lc}
\toprule
 & Raw \\
\midrule
Observed $\rho$   & $+0.780$ \\
95\% CI on $\rho$ & $[+0.368,\,+0.955]$ \\
Null mean $\rho$  & $-0.011 \pm 0.179$ \\
$p(\rho)$         & $<0.005$ \\
\bottomrule
\end{tabular}
\caption{Bootstrap 95\% CI (500 resamples) and permutation test (200 shuffles)
for Channel~2 GB model (Llama-3.2-1B).}
\label{tab:ch2_perm_llama}
\end{table}

\begin{figure*}[htbp]
\centering
\begin{minipage}{0.3\textwidth}
    \centering
    \includegraphics[width=\linewidth]{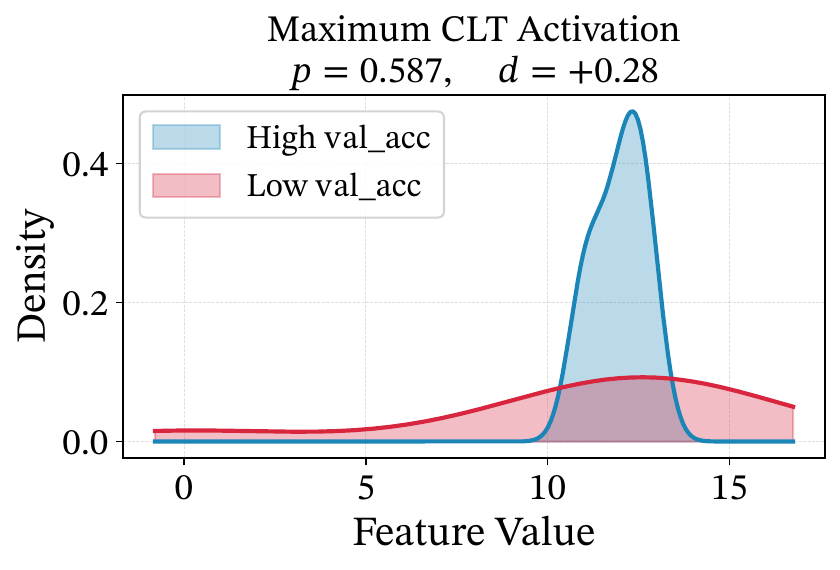}
\end{minipage}\hfill
\begin{minipage}{0.3\textwidth}
    \centering
    \includegraphics[width=\linewidth]{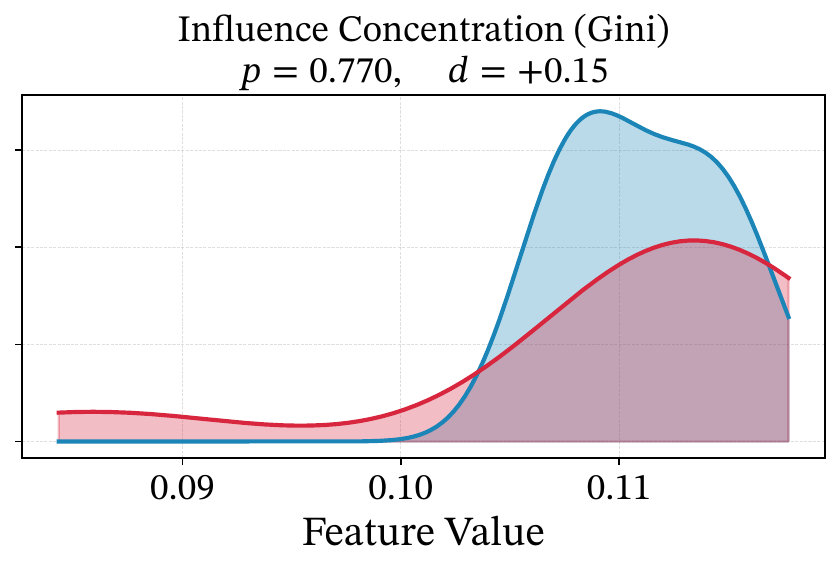}
\end{minipage}\hfill
\begin{minipage}{0.3\textwidth}
    \centering
    \includegraphics[width=\linewidth]{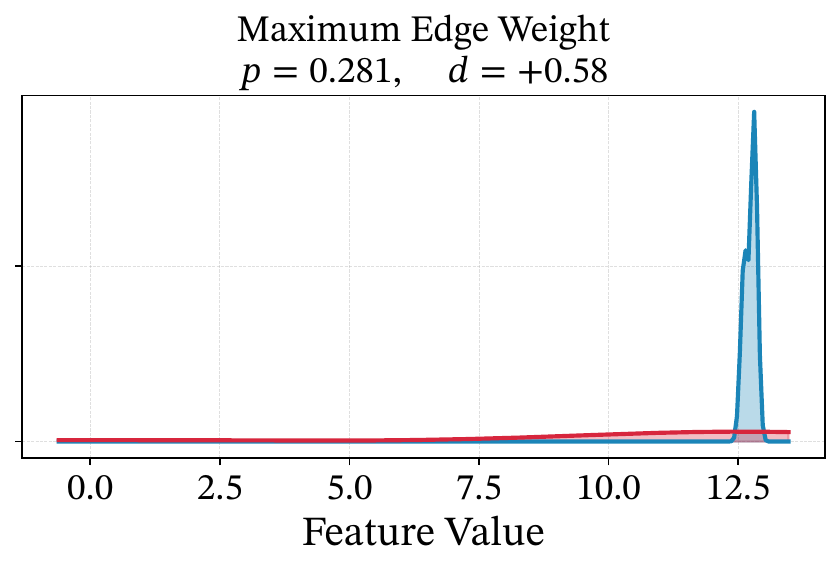}
\end{minipage}\hfill
\begin{minipage}{0.3\textwidth}
    \centering
    \includegraphics[width=\linewidth]{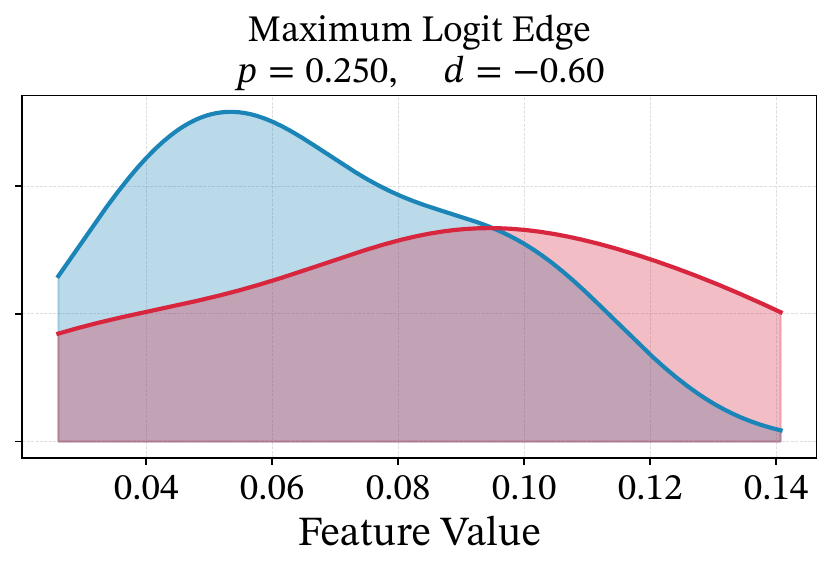}
\end{minipage}
\begin{minipage}{0.3\textwidth}
    \centering
    \includegraphics[width=\linewidth]{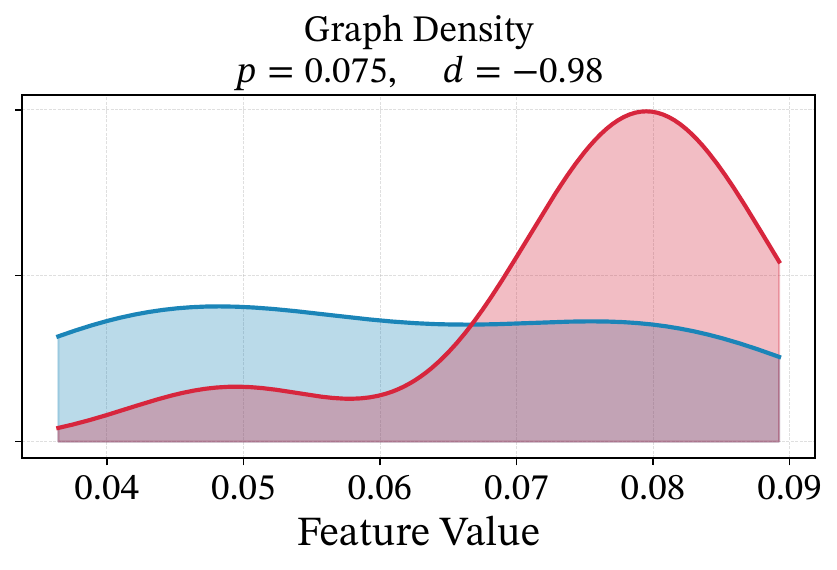}
\end{minipage}
\caption{\textbf{Llama-3.2-1B}: Distributions of five selected graph-structural features for high- versus low-accuracy probe layers, shown for Sentiment. No feature reaches significance at $p < 0.05$, and effect sizes are small, with Cohen's $d$ ranging from $-0.98$ to $+0.58$, consistent with the weak within-concept structural separation observed for Sentiment in Llama-3.2-1B.}
\label{fig:tp_sentiment_llama}
\end{figure*}

\begin{figure*}[htbp]
\centering
\begin{minipage}{0.3\textwidth}
    \centering
    \includegraphics[width=\linewidth]{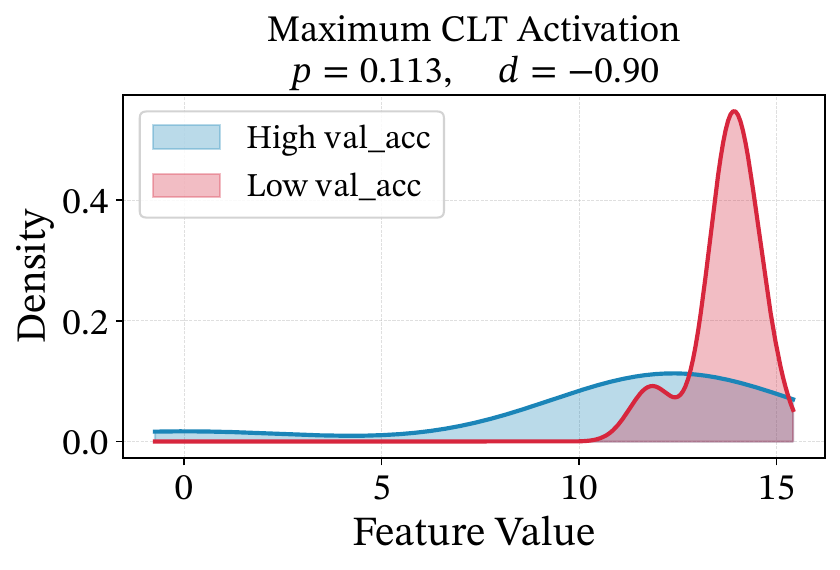}
\end{minipage}\hfill
\begin{minipage}{0.3\textwidth}
    \centering
    \includegraphics[width=\linewidth]{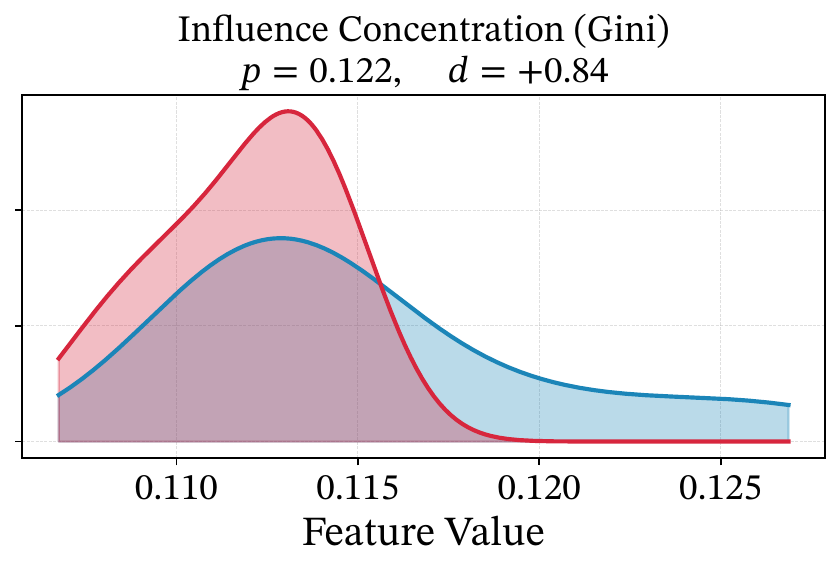}
\end{minipage}\hfill
\begin{minipage}{0.3\textwidth}
    \centering
    \includegraphics[width=\linewidth]{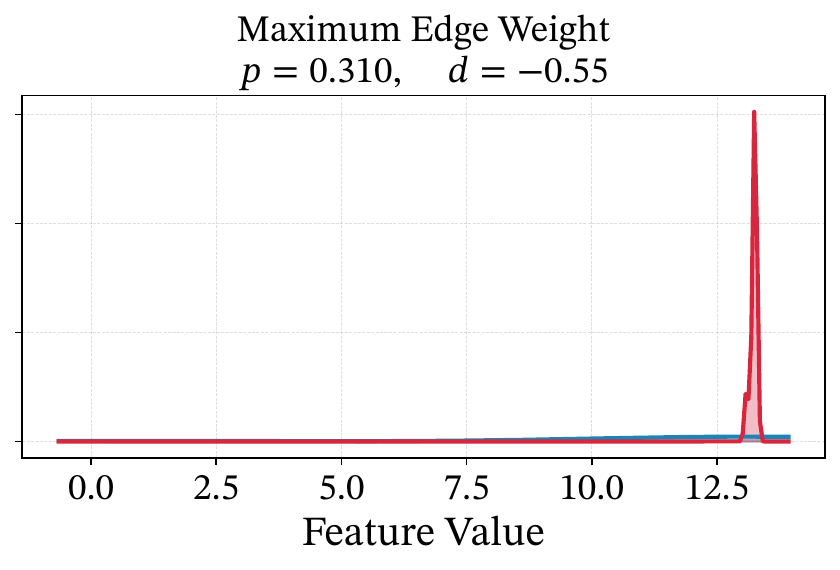}
\end{minipage}\hfill
\begin{minipage}{0.3\textwidth}
    \centering
    \includegraphics[width=\linewidth]{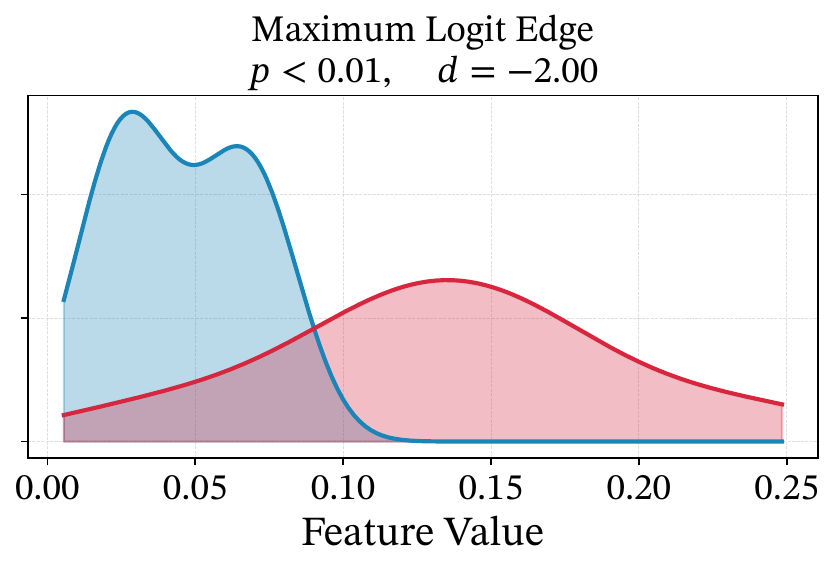}
\end{minipage}
\begin{minipage}{0.3\textwidth}
    \centering
    \includegraphics[width=\linewidth]{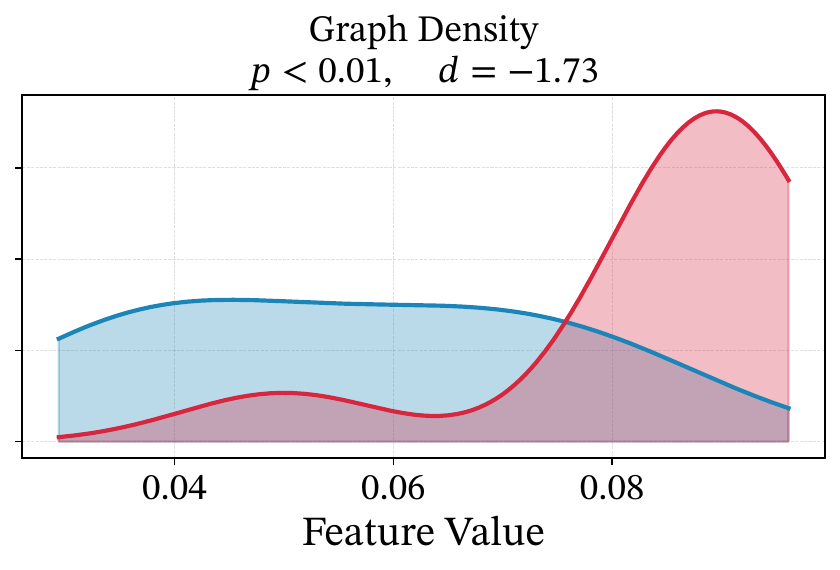}
\end{minipage}
\caption{\textbf{Llama-3.2-1B}: Distributions of five selected graph-structural features for high- versus low-accuracy probe layers, shown for Reasoning. Maximum logit edge and graph density reach significance ($p < 0.005$), with large effect sizes ($d = -2.00$ and $d = -1.73$ respectively); the remaining features do not. Cohen's $d$ ranges from $-2.00$ to $+0.84$. Negative $d$ values indicate that lower-accuracy layers exhibit higher feature values, a pattern inverted relative to Gemma-2-2B.}
\label{fig:tp_reasoning_llama}
\end{figure*}

\begin{figure*}[htbp]
\centering
\begin{minipage}{0.3\textwidth}
    \centering
    \includegraphics[width=\linewidth]{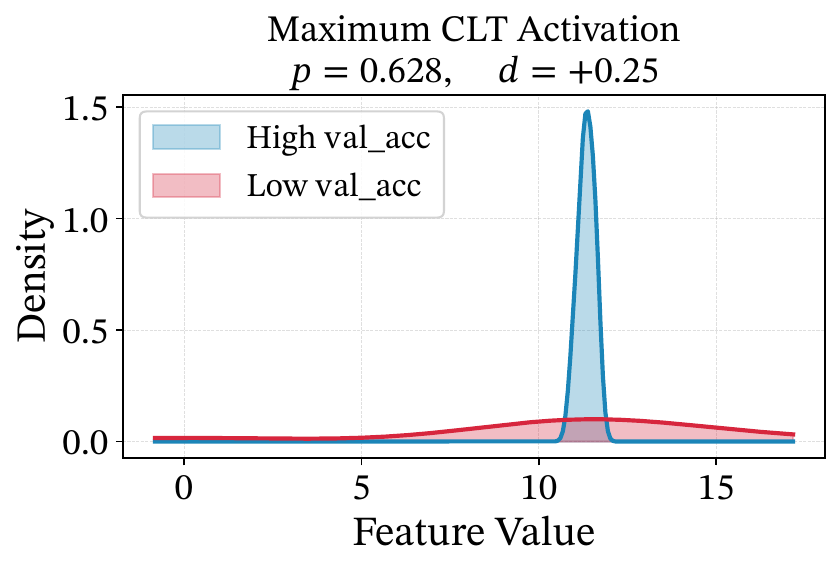}
\end{minipage}\hfill
\begin{minipage}{0.3\textwidth}
    \centering
    \includegraphics[width=\linewidth]{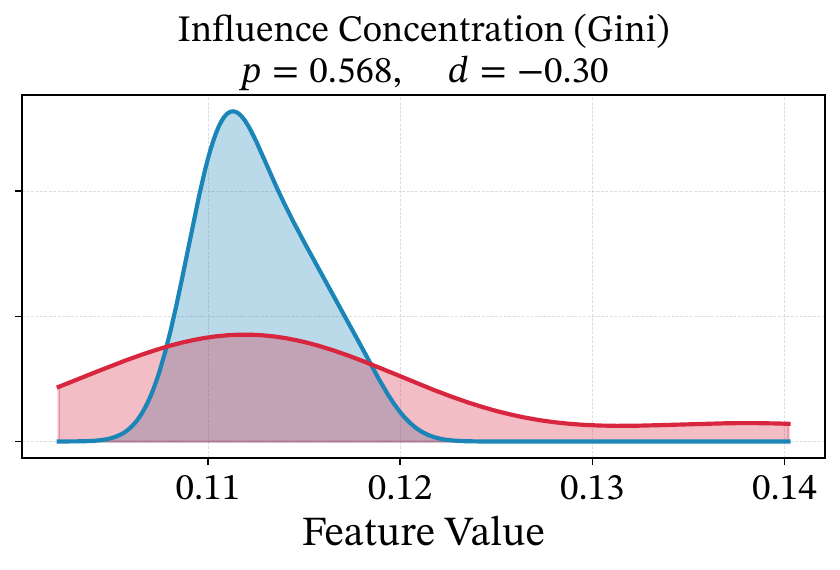}
\end{minipage}\hfill
\begin{minipage}{0.3\textwidth}
    \centering
    \includegraphics[width=\linewidth]{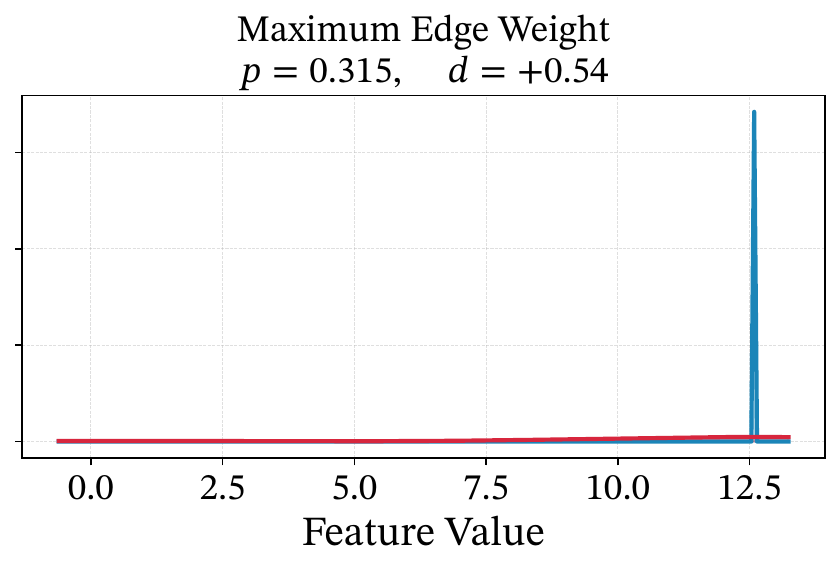}
\end{minipage}\hfill
\begin{minipage}{0.3\textwidth}
    \centering
    \includegraphics[width=\linewidth]{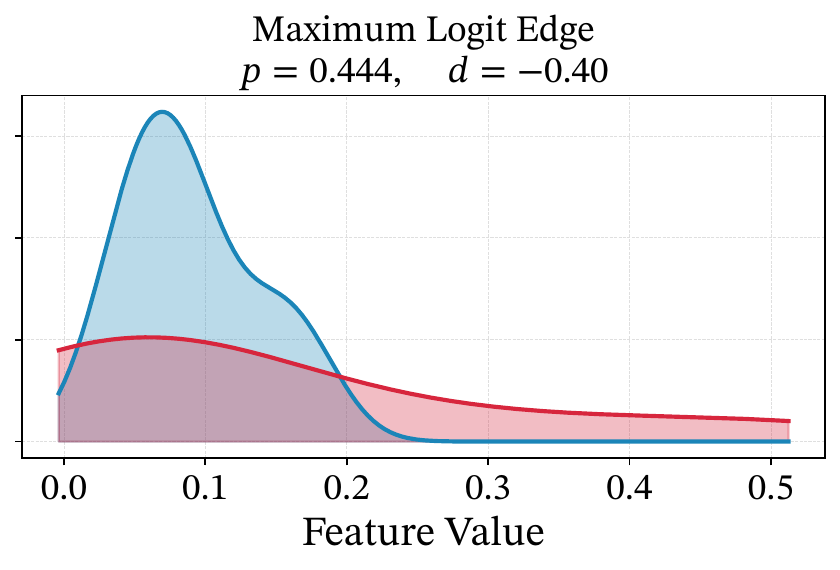}
\end{minipage}
\begin{minipage}{0.3\textwidth}
    \centering
    \includegraphics[width=\linewidth]{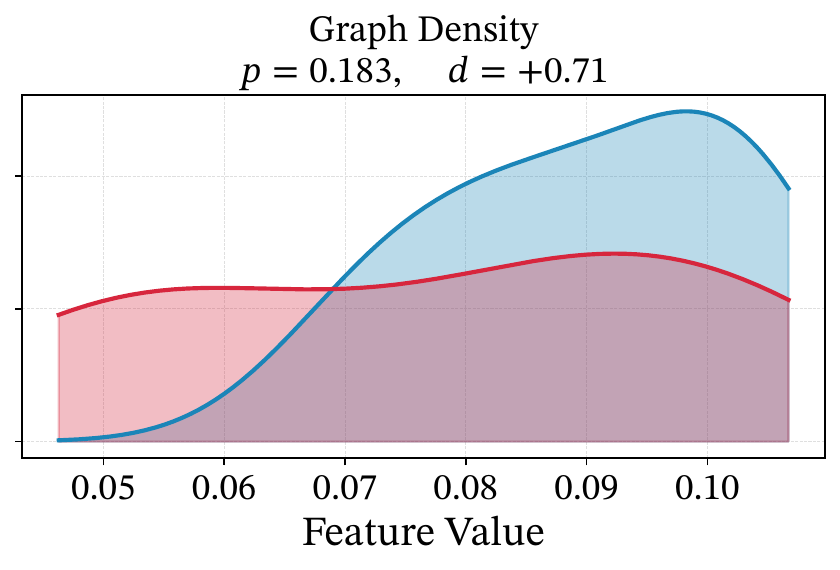}
\end{minipage}
\caption{\textbf{Llama-3.2-1B}: Distributions of five selected graph-structural features for high- versus low-accuracy probe layers, shown for Truthfulness. No feature reaches significance at $p < 0.05$, and effect sizes are small, with Cohen's $d$ ranging from $-0.40$ to $+0.71$, reflecting the near-flat val\_acc trajectory and weak concept encoding of Truthfulness at the 1B scale.}
\label{fig:tp_truthfulness_llama}
\end{figure*}

\noindent\textbf{Structural separation is weaker than Gemma-2-2B.}
While the raw signal is significant, the within-concept (de-meaned) signal
is substantially weaker: $\rho = +0.34$ with a bootstrap CI that straddles
zero $[-0.055, +0.743]$, compared to $\rho = +0.908$ $[+0.817, +0.962]$ for
Gemma-2-2B. Two compounding factors explain this. First, Llama-3.2-1B has
only 16 layers, compressing the structural variation that Channel~2 exploits
across depth. Second, the val\_acc trajectories for Sentiment and Truthfulness
are near-flat across layers (Figure~\ref{fig:probe_accuracy_trajectories}),
leaving little within-concept variance for the regressor to fit. The
topological fingerprint distributions (Figures~\ref{fig:tp_reasoning_llama},
\ref{fig:tp_sentiment_llama}, \ref{fig:tp_reasoning_llama}, \ref{fig:tp_truthfulness_llama})
confirm this: Toxicity and Reasoning show clear separation between high- and
low-accuracy layers on global features such as graph density and mean
indegree, while Sentiment and Truthfulness show little separation on any
feature family.

Concept-transfer is also poor across the board: held-out concept $\rho$
ranges from $-0.74$ to $+0.48$ for the raw target, compared to $+0.93$--$+0.95$
for Sentiment and Reasoning in Gemma-2-2B. With only 16 layers and compressed
structural differences between concepts, the model cannot reliably generalise
the structural signature of a ``good probe layer'' to unseen concepts. Late-layer
holdout (train on $L_0$--$L_7$, test on $L_8$--$L_{15}$) is the exception,
achieving $\rho = +0.90$ and $R^2 = +0.905$, because the dominant signal
(\texttt{n\_embed\_nodes}) is consistent across depth at this model scale.

\newpage
\section{Computational Resources}
All experiments were run on a single NVIDIA A100 GPU (80 GB VRAM). The primary computational cost is attribution graph construction: for each of the four concept categories, we build graphs at every layer for 100 prompts per dataset, yielding approximately 10,400 graphs in total across concepts and layers for Gemma-2-2B. Moreover, for causal dissociation we utilized 50 prompts per concept. The feature-level predictability experiments were the most expensive component at approximately 26 hours, followed by the graph structural predictability analysis at 5 hours, causal ablations at 3 hours, and probe training at 3 hours. The Llama-3.2-1B replication required approximately half the total runtime, owing to the smaller model and shallower layer count.
\newpage
\section{Topological Fingerprints of Graph Structural Features}
\begin{figure*}[ht]
\centering
\begin{minipage}{0.3\textwidth}
    \centering
    \includegraphics[width=\linewidth]{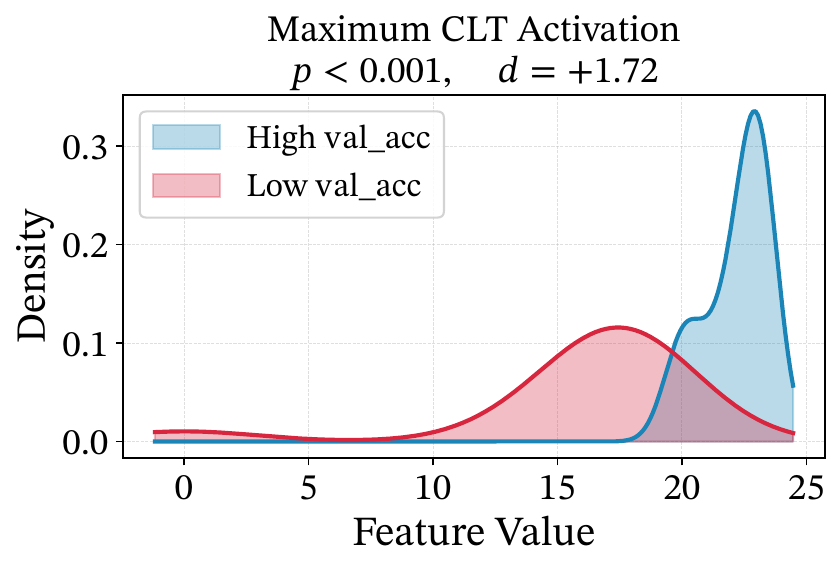}
\end{minipage}\hfill
\begin{minipage}{0.3\textwidth}
    \centering
    \includegraphics[width=\linewidth]{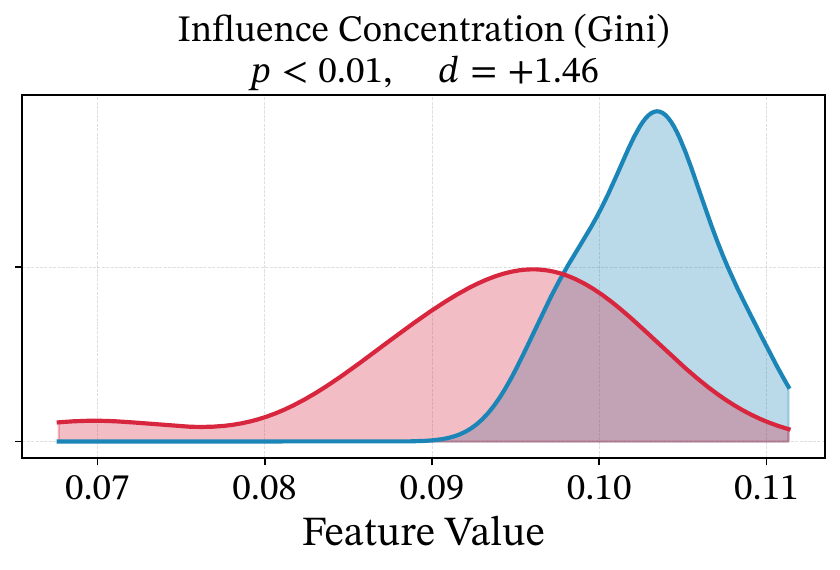}
\end{minipage}\hfill
\begin{minipage}{0.3\textwidth}
    \centering
    \includegraphics[width=\linewidth]{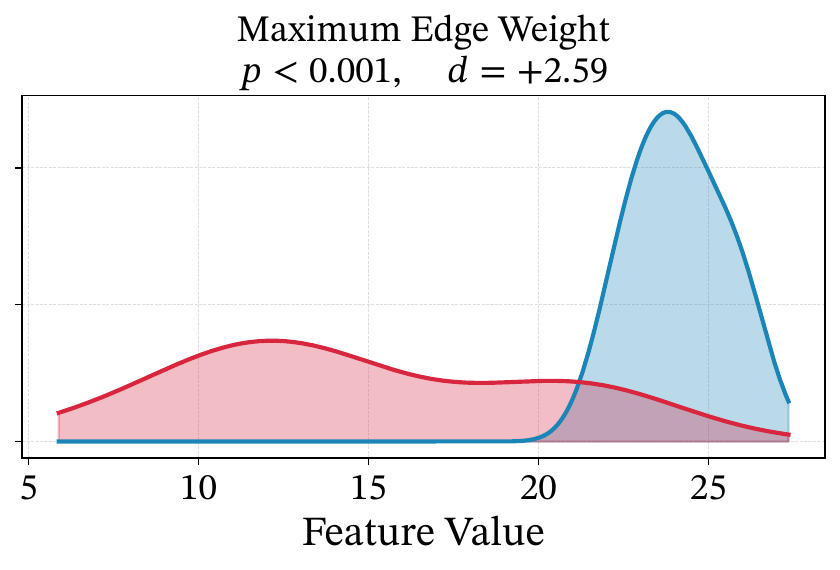}
\end{minipage}\hfill
\begin{minipage}{0.3\textwidth}
    \centering
    \includegraphics[width=\linewidth]{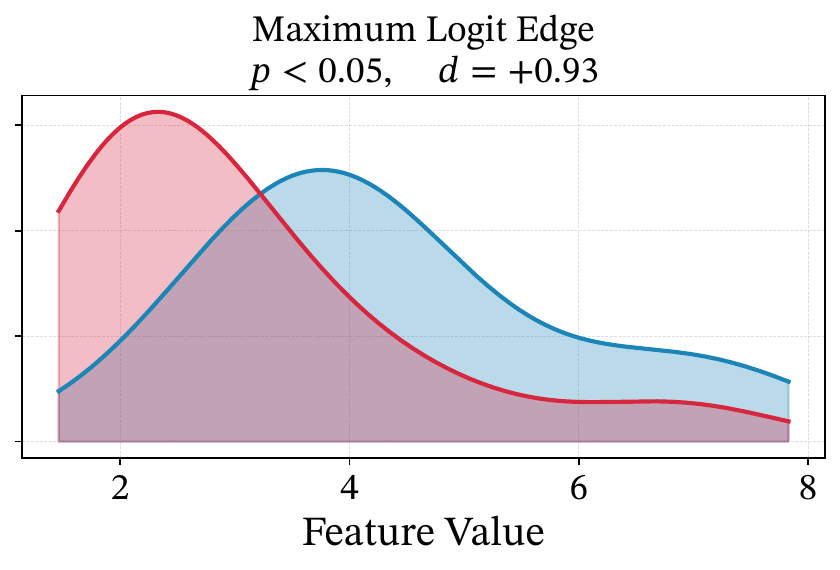}
\end{minipage}
\begin{minipage}{0.3\textwidth}
    \centering
    \includegraphics[width=\linewidth]{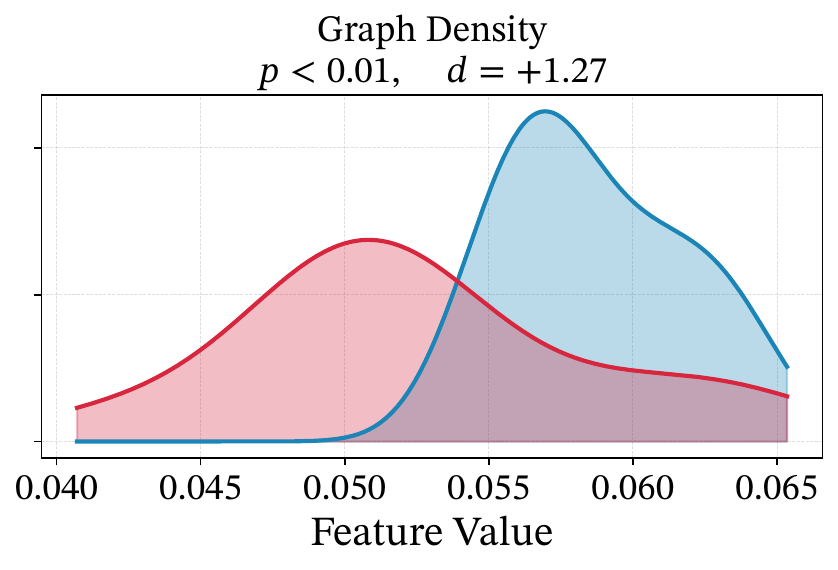}
\end{minipage}
\caption{\textbf{Gemma-2-2B}: Distributions of five selected graph-structural features for high- versus low-accuracy probe layers, shown for sentiment. The separation is statistically significant for each feature ($p < 0.05$, independent $t$-test) and corresponds to medium-to-large effect sizes, with Cohen's $d$ ranging from $+0.93$ to $+2.59$. These results indicate that CTA graphs contain a separable structural signal of a layer's concept-encoding quality.}
\label{fig:topological_fingerprints_sentiment}
\end{figure*}
\begin{figure*}[htbp]
\centering
\begin{minipage}{0.3\textwidth}
    \centering
    \includegraphics[width=\linewidth]{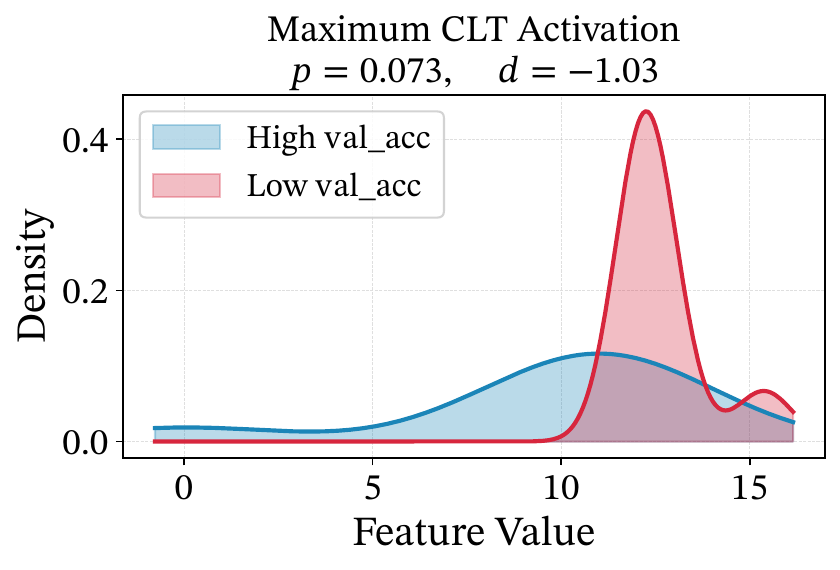}
\end{minipage}\hfill
\begin{minipage}{0.3\textwidth}
    \centering
    \includegraphics[width=\linewidth]{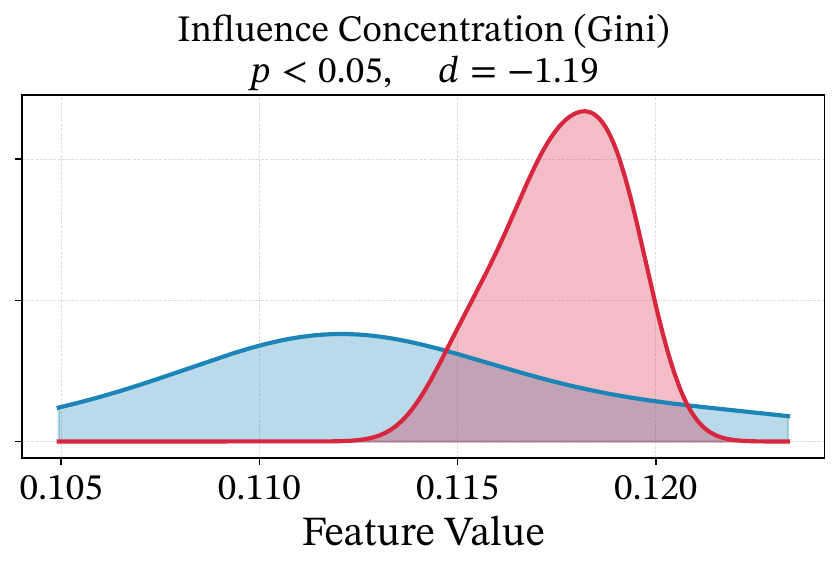}
\end{minipage}\hfill
\begin{minipage}{0.3\textwidth}
    \centering
    \includegraphics[width=\linewidth]{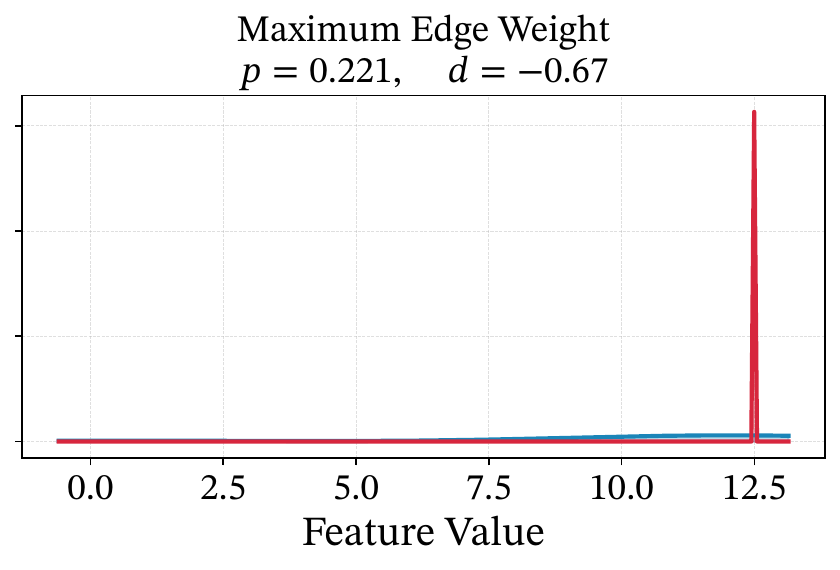}
\end{minipage}\hfill
\begin{minipage}{0.3\textwidth}
    \centering
    \includegraphics[width=\linewidth]{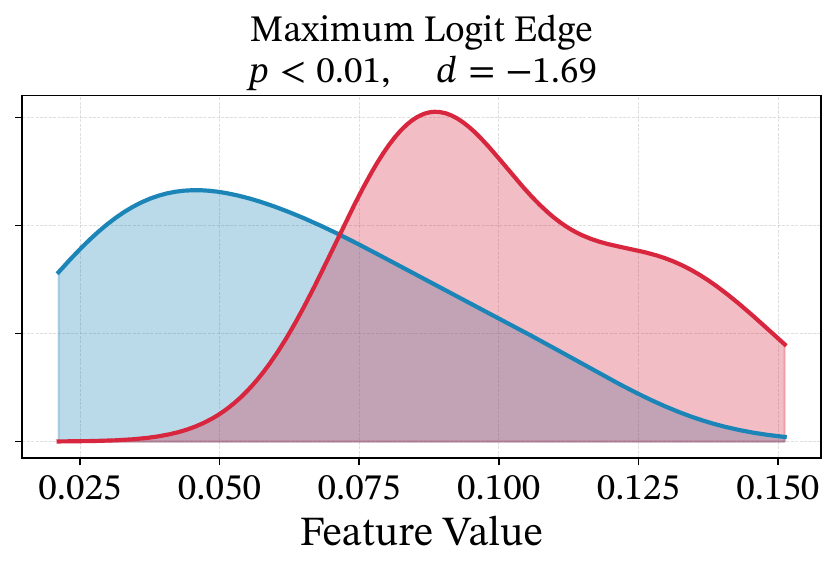}
\end{minipage}
\begin{minipage}{0.3\textwidth}
    \centering
    \includegraphics[width=\linewidth]{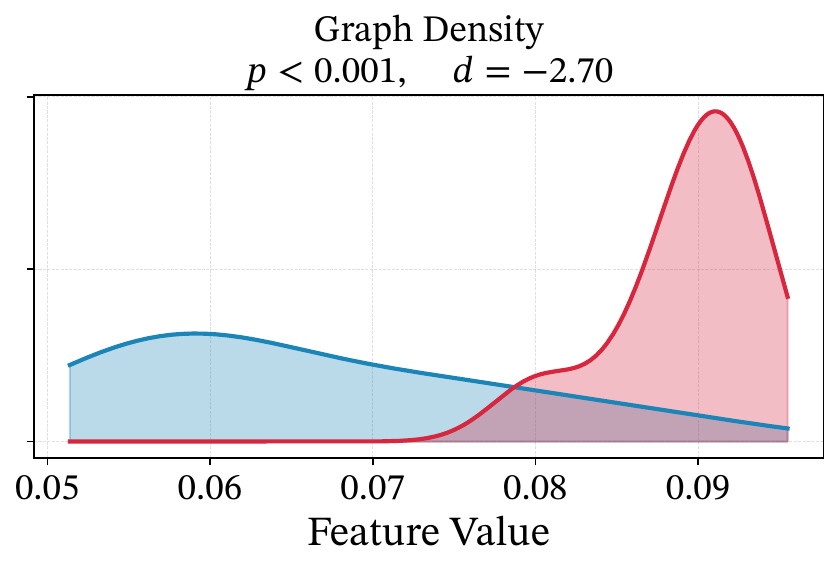}
\end{minipage}
\caption{\textbf{Llama-3.2-1B}: Distributions of five selected graph-structural features for high- versus low-accuracy probe layers, shown for Sentiment. No feature reaches significance at $p < 0.05$, and effect sizes are small, with Cohen's $d$ ranging from $-0.98$ to $+0.58$, consistent with the weak within-concept structural separation observed for Sentiment in Llama-3.2-1B.}
\label{fig:tp_toxicity_llama}
\end{figure*}

% \section{Example Appendix}
% \label{sec:appendix}

% This is an appendix.

\end{document}